\documentclass[letterpaper,journal]{IEEEtran}
\usepackage[T1]{fontenc}
\usepackage[utf8]{inputenc}
\usepackage{amsmath,amsfonts}
\usepackage{graphicx}
\usepackage{booktabs}
\usepackage{url}
\usepackage{array}
\usepackage{multirow}
\usepackage{listings}
\usepackage{cite}
\usepackage{xcolor}
\usepackage{mdframed}
\usepackage[hidelinks]{hyperref}
\newmdenv[linewidth=0.4pt,linecolor=black!60,backgroundcolor=white,
  frametitlebackgroundcolor=black!8,frametitlefont=\normalfont\footnotesize\bfseries,
  frametitlerule=true,frametitlerulewidth=0.4pt,
  frametitleaboveskip=3pt,frametitlebelowskip=3pt,
  frametitleleftmargin=4pt,frametitlerightmargin=4pt,
  innerleftmargin=2pt,innerrightmargin=2pt,innertopmargin=3pt,innerbottommargin=3pt,
  skipabove=3pt,skipbelow=3pt,splittopskip=3pt,splitbottomskip=3pt,
  needspace=4\baselineskip,repeatframetitle=true]{codepanel}
\graphicspath{{figures/}}
\title{RS-Claw-Evolution: Environment-Feedback-Driven Evolution for Lightweight Remote Sensing Agents in Long-Horizon Tasks}
\author{Kai Ouyang, Dongyang Hou, Liangtian Liu, Zeyuan Wang, Ziyu Li,
Chengfu Liu, Zichao Tang, Xuezhi Cui, Shengwu Ouyang, Wentao Yang,
Hanwen Yu, and
Haifeng Li%
\thanks{Kai Ouyang, Dongyang Hou, Liangtian Liu, Zeyuan Wang, Ziyu Li, Chengfu Liu, Zichao Tang, Xuezhi Cui, and Shengwu Ouyang are with the School of Geosciences and Info-Physics, Central South University, Changsha 410083, China.}%
\thanks{Wentao Yang is with the School of Earth Sciences and Spatial Information Engineering, Hunan University of Science and Technology, Xiangtan 411201, China, and also with Sanya Institute, Hunan University of Science and Technology, Sanya 572024, China.}%
\thanks{Hanwen Yu is with the School of Resources and Environment, University of Electronic Science and Technology of China, Chengdu 611700, China.}%
\thanks{Haifeng Li is with the School of Artificial Intelligence, Central South University, Changsha 410083, China.}}
\begin{document}
\maketitle

\begin{abstract}
Large language model-driven remote sensing (RS) agents hold significant potential for automating complex geospatial analysis. However, lightweight RS agents based on compact language models currently struggle with multi-step interactive tasks due to loss of long-horizon states, inefficient environmental feedback utilization, and sparse optimization signals. To address these limitations, this paper proposes RS-Claw-Evolution, an environment-feedback-driven evolution framework for lightweight remote sensing agents in long-horizon tasks, which transforms environmental feedback into effective learning signals for progressive agent capability improvement.
The proposed framework realizes agent evolution through three progressive stages. First, at the interaction evolution stage, a programming-based interaction paradigm is introduced to explicitly control observations and maintain intermediate states via executable code, mitigating context redundancy in long-horizon execution. Second, at the experience evolution stage, to overcome the scarcity of high-quality RS interaction data, we design a feedback-driven supervised fine-tuning strategy. It employs failure-aware trajectory generation to capture informative error patterns, coupled with an error-turn masking mechanism that preserves recovery behaviors without imitating faulty actions. Finally, at the decision evolution stage, a reinforcement learning phase utilizes multi-dimensional environment rewards and turn-level advantage protection to optimize tool-use behaviors and alleviate credit assignment issues in long sequences.
Experiments on the Earth-Bench benchmark quantitatively confirm the effectiveness of RS-Claw-Evolution. Notably, the optimized Qwen3-4B-based agent achieves an accuracy of 65.9\% in Autonomous Planning (AP) mode, substantially outperforming the untrained Qwen3-32B baseline (43.8\%) and surpassing large-scale general-purpose models like DeepSeek-V3.1 (60.8\%), while closely approaching GPT-5 (71.6\%). These results demonstrate that effective environment-feedback integration enables the capability evolution of lightweight agents, narrowing the performance gap with large-scale counterparts in long-horizon remote sensing tasks.
\end{abstract}

\section{Introduction}

\IEEEPARstart{D}{riven} by the reasoning and planning capabilities of large language models (LLMs), remote sensing (RS) agents have emerged as a promising paradigm for automating complex Earth observation tasks~\cite{feng2026earthagent,xu2024rsagent}. Distinct from traditional models constrained to isolated perception tasks, these agents can autonomously orchestrate long-horizon workflows. By dynamically invoking external tools and interacting with the environment, they seamlessly bridge data retrieval, spatial analysis, and result generation~\cite{feng2026earthagent,shabbir2026thinkgeo}. As tool-augmented architectures mature, the capacity to operate across vast toolsets has become a hallmark of advanced agent systems~\cite{qin2024toollm}. Nevertheless, deploying these agents in professional RS domains exposes a critical bottleneck. When executing multi-turn interactive tasks within a massive, specialized tool space, the efficient management of tool representations and interaction contexts becomes the decisive factor for system reliability~\cite{qin2024toollm,liu2026rsclaw}. Conventional approaches typically force all tool descriptions into the model's context window. While straightforward, this brute-force strategy falters in large-scale scenarios. It not only incurs exorbitant context overhead but also triggers semantic interference among tools with overlapping functionalities, substantially elevating the risk of incorrect tool selection and parameter configuration failures~\cite{qin2024toollm,liu2026rsclaw}.

To alleviate context overload, recent frameworks like RS-Claw~\cite{liu2026rsclaw} have restructured flat tool spaces into hierarchical skill trees, enabling agents to dynamically navigate and retrieve tools on demand. While effective at optimizing retrieval, this architectural workaround heavily relies on the base language model's innate ability to interpret execution results and environmental feedback. This dependency becomes a critical vulnerability when scaling down to parameter-constrained lightweight models. During long-horizon interactions, complex execution outputs, runtime exceptions, and dynamic environmental feedback can introduce out-of-distribution (OOD) interaction contexts~\cite{gudibande2024falsepromise}, whose underlying states and transitions are poorly covered by conventional LLM training data~\cite{chen2023fireact,zeng2024agenttuning}. Consequently, lightweight agents struggle to maintain stable state representations, frequently deteriorating into redundant invocations, error loops, and eventual task failures~\cite{feng2026earthagent,shinn2023reflexion,shabbir2026thinkgeo}. These phenomena reveal that structural organization alone, while reducing environmental complexity, cannot fundamentally compensate for a lightweight model's intrinsic reasoning deficits. To achieve genuine autonomy, it is imperative to move beyond inference-time prompting and directly internalize tool invocation logic, continuous state tracking, and resilient error recovery into the model parameters via targeted training. Nevertheless, realizing this feedback-driven training paradigm in the RS domain confronts three formidable challenges.

\begin{figure*}[!t]
\centering
\includegraphics[width=\linewidth]{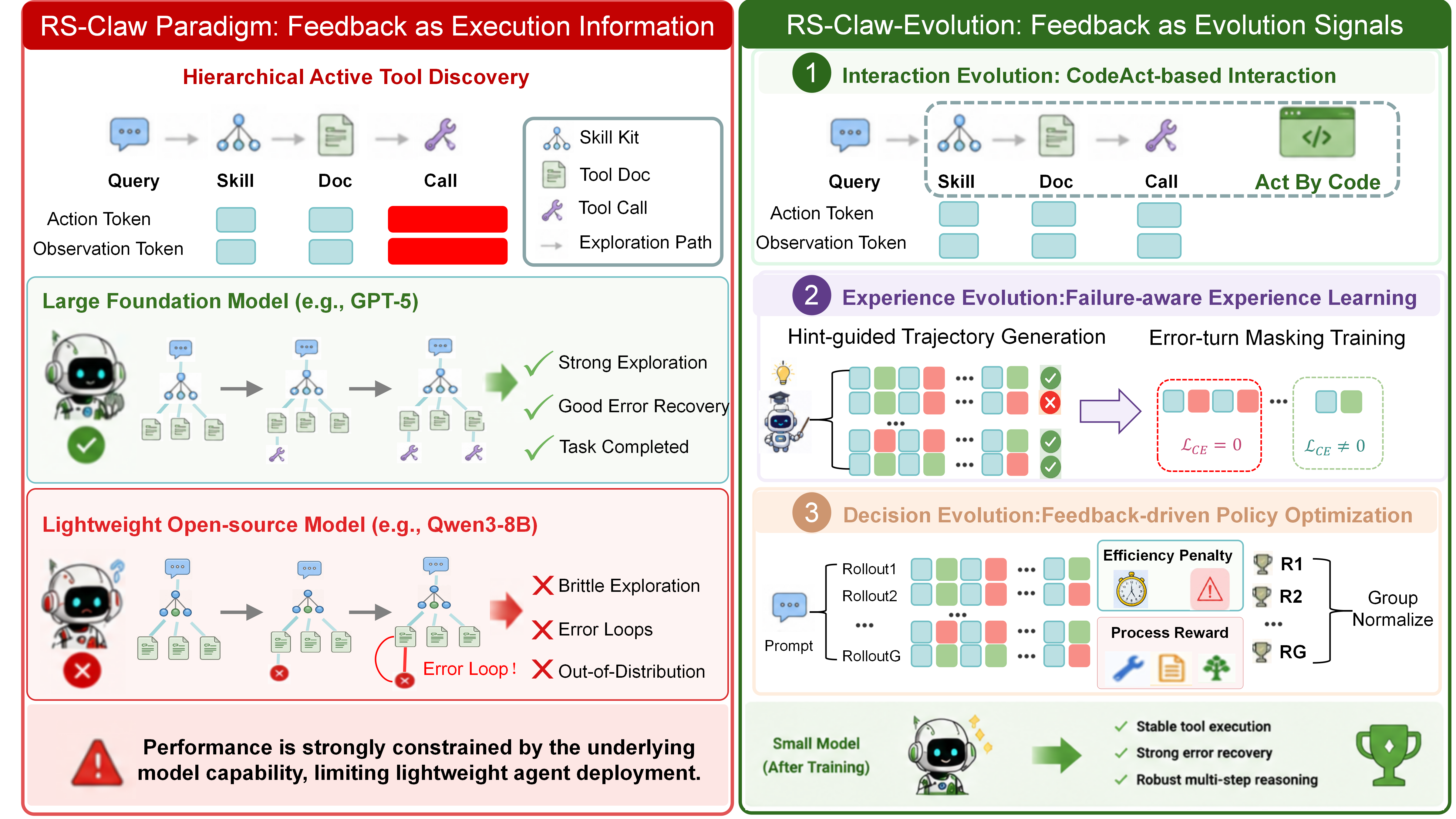}
\caption{Comparison between the original RS-Claw paradigm and the proposed RS-Claw-Evolution framework. The original RS-Claw paradigm treats environmental feedback as execution information during inference, while RS-Claw-Evolution further exploits environmental feedback as evolution signals to progressively improve lightweight agents through three stages: Interaction Evolution via CodeAct-based interaction, Experience Evolution via failure-aware experience learning, and Decision Evolution via feedback-driven policy optimization.}
\label{fig:framework}
\end{figure*}

\textbf{1) Context Inflation in Long-Horizon Workflows:} RS tasks inherently involve multi-stage processing pipelines that continuously generate intermediate artifacts, such as file paths, parameter configurations, and execution logs~\cite{feng2026earthagent,liu2026rsclaw,kang2026acon}. As interaction turns multiply, the uncontrolled accumulation of these environmental states imposes prohibitive memory constraints during long-trajectory training. Developing mechanisms to distil redundant interaction histories while preserving critical environmental integrity remains a fundamental prerequisite for efficient agent training.

\textbf{2) The Scarcity of High-Fidelity Data and the Success-Bias Trap:} Unlike general-purpose agents that leverage scalable simulated environments such as WebArena and OSWorld to automatically synthesize massive datasets~\cite{zhou2024webarena,xie2024osworld}, RS agents are strictly constrained by real-world geospatial data, rigid file formats, and domain-specific attributes. This lack of scalable simulation severely bottlenecks task generation. Consequently, current paradigms~\cite{chen2023fireact,zeng2024agenttuning} often rely on costly distillation from powerful proprietary models to construct perfect demonstration trajectories. However, training parameter-constrained models exclusively on these idealized, error-free trajectories leads to severe overfitting. Bereft of exposure to erroneous exploration, abnormal feedback, and subsequent recovery behaviors, lightweight agents fail drastically when encountering real-world tool failures. Thus, under conditions of restricted environment scalability, maximizing the instructional value of limited, imperfect interaction trajectories is crucial for robust generalization.

\textbf{3) Sparse Optimization Signals and Underutilized Environmental Feedback:} After supervised fine-tuning (SFT) establishes basic tool invocation capabilities, reinforcement learning (RL) can further optimize agent policies. Proximal Policy Optimization (PPO)~\cite{schulman2017ppo} and Group Relative Policy Optimization (GRPO)~\cite{shao2024deepseekmath} perform reward-based policy optimization, whereas Direct Preference Optimization (DPO)~\cite{rafailov2023dpo} directly learns from preference pairs. When these methods are configured with terminal task rewards or whole-trajectory preferences, their supervision may inadequately distinguish intermediate contributions in long-horizon tool interactions. Process supervision~\cite{lightman2024verify,wang2024mathshepherd} and agent-specific credit assignment methods~\cite{luo2025agentlightning,feng2025gigpo} address this limitation through different mechanisms. Learned evaluators require additional training or inference, whereas Group-in-Group Policy Optimization (GiGPO) derives step-level advantages through repeated-state grouping without auxiliary models or additional rollouts. These approaches motivate a complementary question for RS agents: how can native execution states, runtime exceptions, and task progress be used to construct informative optimization signals without an auxiliary process evaluation model?

Motivated by these challenges, this paper re-examines the learning paradigm of lightweight RS agents from the perspective of continuous agent-environment interaction. We argue that environmental feedback should not merely serve as an inference-time signal for updating agent states, but should be progressively transformed into training-time supervision to drive agent capability evolution. As illustrated in Fig.~\ref{fig:framework}, we propose RS-Claw-Evolution, an environment-feedback-driven evolution framework that improves lightweight RS agents through three progressive stages: Interaction Evolution, Experience Evolution, and Decision Evolution. First, at the interaction evolution stage, we introduce a programming-based interaction paradigm that adopts executable code as the action representation. By delegating intermediate state management and execution logic to the environment, this paradigm reduces redundant context accumulation and establishes an efficient interaction foundation for long-horizon tasks. Second, at the experience evolution stage, we develop a feedback-driven SFT strategy that incorporates failure feedback and recovery trajectories into agent training. Instead of relying solely on idealized successful demonstrations, this strategy enables lightweight agents to acquire richer interaction experience while avoiding the imitation of erroneous behaviors. Finally, at the decision evolution stage, we transform environmental feedback into dense optimization signals for RL. These signals support finer-grained credit assignment and guide tool-use behaviors, helping agents make more reliable and efficient decisions in long-horizon tasks.

The main contributions of this paper are summarized as follows:

\textbf{1) A novel feedback-driven evolution framework:} We introduce RS-Claw-Evolution, a comprehensive training framework for lightweight RS agents that progressively enhances agent capabilities through three evolutionary stages: Interaction Evolution, Experience Evolution, and Decision Evolution. By transforming environmental feedback from passive inference-time observations into active training-time optimization signals, the proposed framework enables continuous capability improvement for long-horizon RS tasks.

\textbf{2) Experience evolution through failure-aware SFT:} To overcome the scarcity of high-quality RS interaction data and the high cost of expert trajectory construction, we propose a hint-guided failure-aware trajectory generation method. It leverages post-hoc analysis of failed explorations to extract task-level guidance and efficiently generate informative trajectories containing recovery processes. Coupled with the Error-turn Masking mechanism, this strategy prevents the model from imitating erroneous actions while preserving valuable error feedback and correction behaviors.

\textbf{3) Decision evolution through feedback-driven RL:} To address sparse optimization signals and complex credit assignment in long-horizon tasks, we formulate a multi-dimensional environment feedback reward (MEFR) that incorporates tool execution states, document comprehension, data flow integrity, and dynamic penalties. Furthermore, we design a turn-level advantage protection (TAP) mechanism to suppress erroneous decisions while preserving effective recovery behaviors during policy optimization.

\textbf{4) Extensive empirical validation:} We conduct systematic experiments on the comprehensive RS agent benchmark, Earth-Bench~\cite{feng2026earthagent}. The results demonstrate that RS-Claw-Evolution enables lightweight models to achieve highly competitive performance in long-horizon autonomous RS tasks. Using only a limited corpus of real RS training data, the proposed framework significantly improves tool invocation, multi-stage task planning, and robust error recovery capabilities.

\section{Related Work}

\subsection{Autonomous Agents for RS}

LLM-based agents organize task execution through a reasoning-action-feedback loop. ReAct~\cite{yao2023react} interleaves language reasoning with external actions, enabling models to dynamically interact with environments, while Reflexion~\cite{shinn2023reflexion} introduces verbal reflection over execution outcomes to improve subsequent attempts. These studies establish environmental feedback as an important resource for inference-time behavior adaptation.

In RS, vision-language models provide an important foundation for semantic understanding and user interaction. RemoteCLIP~\cite{liu2024remoteclip} learns RS image-text representations through large-scale contrastive learning, while RSGPT~\cite{hu2025rsgpt} develops instruction-tuned RS vision-language models for image captioning and visual question answering. GeoChat~\cite{kuckreja2024geochat} introduces region-level inputs and spatial coordinate representations for fine-grained spatial understanding. EarthGPT~\cite{zhang2024earthgpt} integrates multi-source sensor information through instruction tuning, and EarthDial~\cite{soni2025earthdial} further explores interactive dialogue over multisensor Earth observation data. These models strengthen the perception and interpretation of RS data, while scientific workflows additionally require agents to execute multi-stage operations, manage intermediate states, and adapt to runtime feedback.

Recent research has therefore extended RS models toward tool-augmented autonomous agents. RS-Agent~\cite{xu2024rsagent} explores LLM-driven RS agents by integrating task decomposition, knowledge retrieval, and multi-tool scheduling for tasks such as scene classification, visual question answering, and object counting. Earth-Agent~\cite{feng2026earthagent} further constructs a scientific Earth observation agent by integrating 104 domain-specific tools through the Model Context Protocol (MCP) ecosystem, enabling multimodal, multi-step, and quantitative spatio-temporal analysis. Earth-Agent also introduces Earth-Bench, which evaluates both final task outcomes and intermediate agent trajectories, providing a basis for assessing the reliability of RS workflow execution. ChangeAgent~\cite{liu2024changeagent} applies agent-based interaction to RS change analysis by combining temporal image understanding with user-guided reasoning. GeoLLM-Squad~\cite{lee2025geocopilots} explores multi-agent collaboration for decomposing complex geospatial workflows into specialized subtasks. ThinkGeo~\cite{shabbir2026thinkgeo} further studies the evaluation of tool-augmented RS agents, moving the field toward system-level capability assessment. Together, these works establish RS agents as a new paradigm for executing complex Earth observation workflows.

Alongside agent architectures, recent studies have explored improving execution capability through workflow design and model training. Geo-OLM~\cite{stamoulis2025geo} introduces state-driven workflows that decouple task progression from tool invocation, reducing the reasoning burden of geospatial execution without modifying model parameters. OpenEarthAgent~\cite{shabbir2026openearthagent} applies SFT to teach language models verified multi-step geospatial tool interactions. RemoteAgent~\cite{yao2026remoteagent} further investigates reinforcement fine-tuning for improving intent understanding and selective tool invocation in RS tasks. These studies demonstrate that workflow organization, supervised learning, and reinforcement optimization provide complementary approaches for enhancing agent capabilities.

A remaining challenge lies in enabling agents to reliably operate within large-scale heterogeneous tool environments. RS-Claw~\cite{liu2026rsclaw} addresses tool scalability through hierarchical skill trees and progressive information disclosure, allowing agents to retrieve relevant tools without loading all descriptions into the context. However, reducing tool retrieval complexity does not necessarily solve the subsequent execution challenges. Long-horizon RS workflows still involve intermediate files, parameter dependencies, runtime exceptions, and evolving task states that require agents to interpret feedback and adjust future actions. Therefore, reliable tool use depends not only on accessing appropriate capabilities, but also on learning how to interact with environments over extended execution processes.

Overall, existing RS agent research has progressed from visual understanding toward tool-based autonomous analysis, supported by advances in workflow organization, training strategies, and evaluation benchmarks. Nevertheless, lightweight agents still face difficulties in maintaining states, utilizing execution feedback, and recovering from errors during long-horizon interactions. This motivates the investigation of feedback-driven capability evolution, where environmental interactions become not only execution information but also learning signals for improving agent behaviors.

\subsection{Training Long-horizon Tool-use Agents}

Beyond designing agent architectures, recent studies have investigated how language models can acquire reliable tool-use capabilities through training. Toolformer~\cite{schick2023toolformer} explores self-supervised tool-use data construction, enabling models to learn when and how to invoke external tools. FireAct~\cite{chen2023fireact} studies fine-tuning language models with multi-task agent trajectories and demonstrates the importance of agent-specific supervision. AgentTuning~\cite{zeng2024agenttuning} constructs the AgentInstruct dataset and combines agent interaction trajectories with general instruction data during training to improve general agent capabilities. ToolLLM~\cite{qin2024toollm} further scales tool-use training by constructing large-scale application programming interface (API) invocation trajectories, targeting tool selection and parameter generation in complex tool environments. Executable environments and benchmarks, including WebArena~\cite{zhou2024webarena}, SWE-bench~\cite{jimenez2024swebench}, and OSWorld~\cite{xie2024osworld}, provide settings for collecting and evaluating agent interaction trajectories.

The construction of high-quality interaction trajectories is therefore a fundamental factor in agent training. Successful demonstrations provide useful supervision, but may not adequately cover the states that a learned policy encounters after making its own errors. Imitation learning theory shows that behavioral cloning suffers from distribution shift when the learned policy enters states outside the expert distribution~\cite{ross2011imitation}. Studies of model imitation also caution that demonstrations from stronger teachers do not necessarily transfer their broader capabilities to smaller models~\cite{gudibande2024falsepromise}. This issue is particularly relevant to long-horizon tool-use scenarios, where invalid parameters, execution failures, and intermediate state changes may frequently occur. Importantly, failed attempts are not necessarily useless: a successful trajectory may contain incorrect actions followed by effective recovery behaviors. Therefore, understanding how to preserve informative failure experiences while avoiding imitation of invalid actions remains an important challenge.

Another line of research investigates how environmental feedback can improve agent behaviors without directly modifying model parameters. Self-Refine~\cite{madaan2023selfrefine} improves generation through iterative self-feedback, Reflexion~\cite{shinn2023reflexion} converts execution outcomes into textual reflections for subsequent attempts, and CRITIC~\cite{gou2024critic} incorporates external feedback for interactive correction. These approaches demonstrate the value of feedback for improving inference-time behaviors. However, the feedback is primarily used as additional context or memory during execution, rather than being systematically transformed into training signals for improving reusable interaction policies.

Beyond supervised imitation, reward- and preference-based policy optimization provide additional mechanisms for improving agent decision-making. PPO~\cite{schulman2017ppo} introduces clipped policy optimization to stabilize policy updates, while DPO~\cite{rafailov2023dpo} directly learns from preference pairs without explicit reward modeling. GRPO~\cite{shao2024deepseekmath} estimates relative advantages among sampled trajectories without requiring an additional value model, reducing computational overhead. These methods provide different policy optimization frameworks, but the effectiveness of learning depends strongly on the granularity and quality of available supervision signals. When optimization relies only on trajectory-level outcomes, intermediate decisions within long-horizon executions remain difficult to evaluate.

To obtain finer-grained optimization signals, process supervision and learned evaluators have been explored. Lightman et al.~\cite{lightman2024verify} introduce process reward models for mathematical reasoning and release PRM800K with step-level correctness annotations. Math-Shepherd~\cite{wang2024mathshepherd} further explores automatic process supervision construction. JudgeLM~\cite{zhu2025judgelm} studies scalable language-model-based evaluation, while Constitutional AI~\cite{bai2022constitutional} combines self-critique with artificial intelligence (AI)-generated preference feedback for alignment. These approaches illustrate complementary uses of feedback for process supervision, response evaluation, and preference-based alignment. However, applying such evaluation mechanisms to RS agents requires considering domain-specific execution evidence, including file compatibility, data dependencies, tool execution status, and intermediate artifacts. Natural-language evaluation alone cannot fully verify whether an RS operation has actually succeeded.

Recent agent-specific RL methods further investigate trajectory structure and credit assignment. Agent Lightning~\cite{luo2025agentlightning} provides a unified framework for converting agent execution into RL trajectories. GiGPO~\cite{feng2025gigpo} exploits repeated environment states to estimate finer-grained advantages without auxiliary models. Tree-GRPO~\cite{ji2026treegrpo} uses prefix-sharing trajectory structures to derive step-level signals from sampled outcomes. Agentic Reinforced Policy Optimization (ARPO)~\cite{dong2026arpo} explores adaptive rollout strategies and advantage attribution for agent training. These studies highlight the importance of utilizing trajectory structures to improve optimization efficiency in long-horizon agents.

Taken together, existing studies reveal three complementary roles of feedback in agent learning: improving individual attempts during inference, constructing informative experiences for supervised learning, and providing optimization signals for policy improvement. Applying these approaches to RS agents presents challenges arising from limited executable task diversity and the need for execution-grounded evaluation. Assessing intermediate correctness requires evidence from actual tool execution, data dependencies, and workflow progress rather than language-level assessment alone. RS-Claw-Evolution addresses these challenges by transforming environmental feedback into learning signals across supervised learning and reinforcement optimization. It leverages failure-aware experiences to improve error recovery and execution-aware feedback to optimize long-horizon decisions, with the aim of improving tool-use reliability for lightweight agents in heterogeneous RS environments.

\section{Method}

\subsection{Task Formalization}
\label{sec:task-formalization}

We model the multi-turn interaction process between the agent and the RS code execution environment as a discrete-time Partially Observable Markov Decision Process (POMDP), strictly defined by the tuple $\mathcal{M} = \langle \mathcal{Z}, \mathcal{A}, \Omega, \mathcal{T}, \mathcal{R} \rangle$:

\begin{itemize}
\item \textbf{State space }$\mathcal{Z}$\textbf{:} Represents the underlying true physical and logical states of the system, including memory variables in the Python execution environment, underlying multi-source RS datasets, and internal execution states of specialized tools. The global state $s_t \in \mathcal{Z}$ is unobservable to the model.
\item \textbf{Observation space }$\Omega$\textbf{:} Represents the local context information actually received by the model. An observation $o_t \in \Omega$ is the system feedback (e.g., standard output, Error Traceback) disclosed to the model by the environment based on the current hidden state after an action is executed.
\item \textbf{Action space }$\mathcal{A}$\textbf{:} We macroscopically define a single action $a_t \in \mathcal{A}$ as the complete token sequence generated by the model in a single invocation.
\item \textbf{State transition }$\mathcal{T}(s_{t+1}\vert{}s_t, a_t)$\textbf{:} Represents the objective evolution of the environmental state after code execution.
\item \textbf{Reward function }$\mathcal{R}(s_t, a_t)$\textbf{:} Maps state-action pairs to the fine-grained reward $R(\tau)$ we design in the subsequent RL phase.
\end{itemize}

Based on the above framework, we define one complete interaction cycle between the model and the environment as a Turn, strictly corresponding to a single time step $t$ in the POMDP. In the $t$-th turn, the decision state upon which the model relies is solely its experienced historical interaction trajectory $\tau_t$:

\begin{equation}
\tau_t = (o_0, a_1, o_1, a_2, o_2, \dots, a_{t-1}, o_{t-1})
\label{eq:1}
\end{equation}

where $o_0$ is the initial task instruction and base context. In the $t$-th turn, the model samples and generates the complete action sequence $a_t$ for the current turn based on the policy $\pi_\theta(a_t \mid \tau_t)$; subsequently, the environment executes $a_t$, a state transition occurs, and a new observation $o_t$ is returned. The historical trajectory is thus updated to $\tau_{t+1} = (\tau_t, a_t, o_t)$. Specifically, when the agent's action $a_t$ triggers an explicit exception in the underlying environment (e.g., Python syntax error, tool execution crash) or violates task rule constraints (e.g., formatting violation, failure to expand according to the hierarchical skill tree), causing the returned observation $o_t$ to contain an error indicator, we define the error indicator function $\mathcal{E}(a_t, o_t) = 1$. This is used for the subsequently designed error-turn masking and TAP mechanisms (the detailed implementation of this determination is provided in Appendix C of the Supplementary Material). Simultaneously, when the agent generates the terminal action $a_T$ to submit the task result, the interaction process terminates, yielding the complete trajectory $\tau = (o_0, a_1, o_1, \dots, a_{T-1}, o_{T-1}, a_T)$. We extract the trajectory's final predicted answer $\hat{y}_\tau$ from the terminal action $a_T$, where $y^\star$ is the standard answer for the task, used to evaluate whether the model correctly completed it.

\subsection{Interaction Evolution: CodeAct-based Environment Interaction}

Following CodeAct~\cite{wang2024codeact}, we replace the traditional multi-turn interaction paradigm based on JavaScript Object Notation (JSON) with continuous Python code generation to address the complex data flows and lengthy context bottlenecks in RS tasks. In real-world RS analysis workflows, tasks often involve multi-source data fusion and multi-stage processing, generating a massive amount of intermediate products. The storage paths, naming conventions, and metadata of these files are typically extremely lengthy. If the traditional JSON tool invocation mechanism is maintained, the model must passively and repeatedly read and output these massive string constants across multi-turn interactions. This not only rapidly consumes the context window but also creates severe semantic noise, distracting the model from its core reasoning logic.

Unlike tool invocation based on explicit parameter serialization, treating code as action representations allows the agent to implicitly maintain intermediate states through variable reuse, control flow, and program state management. Through programmatic tool invocation, we compress the cumbersome tool execution process into compact logical code. The model can establish compact symbolic references via variable assignments, while the Python execution environment stores the corresponding paths and intermediate values. This reduces repeated transmission of long paths and other verbose state information in the model's context.

More importantly, this design enables the agent to actively control how and at what granularity intermediate information is acquired. The agent can selectively expose intermediate information through the code it writes, while the environment also returns execution exceptions and system feedback. Like human programmers, the model can actively observe critical system feedback through explicit output instructions, thereby avoiding the "passive forced infusion" of environmental feedback into the context seen in traditional frameworks. This mechanism effectively reduces the accumulation of redundant information, stably controlling the context overhead of long-horizon RS tasks within the boundaries of affordable training compute.

\subsection{Experience Evolution: Feedback-driven SFT}

\begin{figure*}[!t]
\centering
\includegraphics[width=\linewidth]{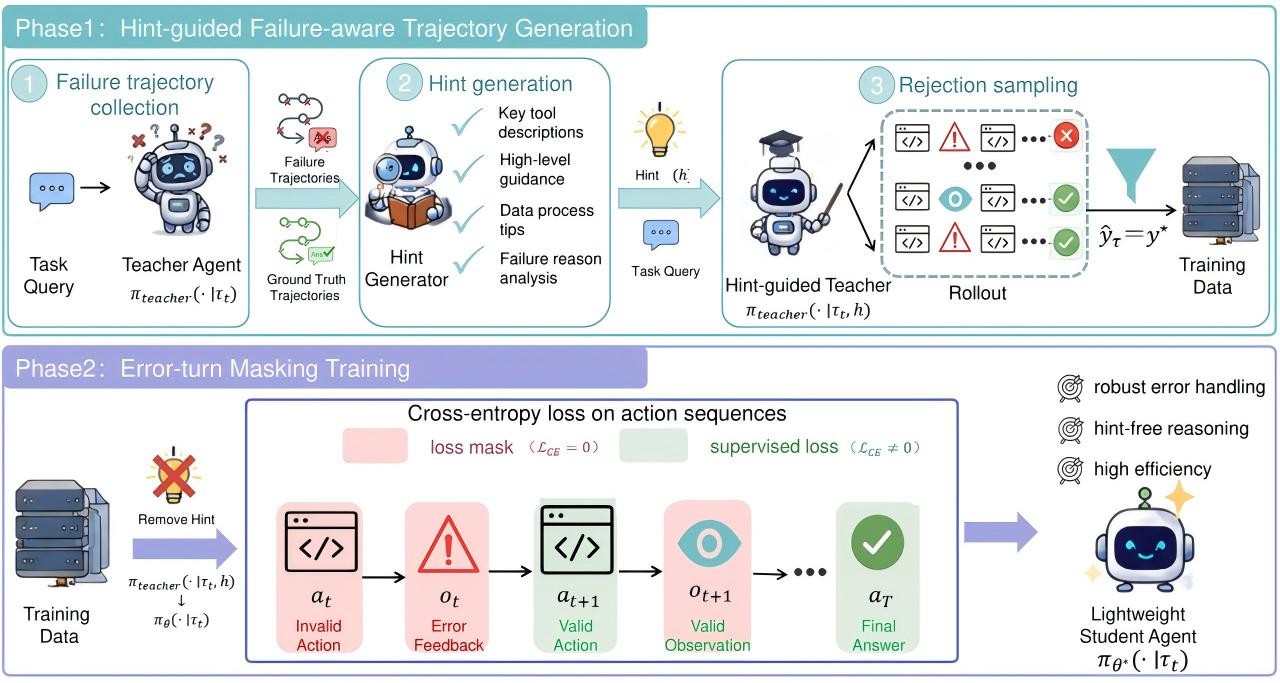}
\caption{Overview of the Experience Evolution stage in RS-Claw-Evolution: feedback-driven SFT for lightweight RS agents. (Top) Hint-guided Failure-aware Trajectory Generation extracts task-level hints through hindsight analysis of failed trajectories with reference to successful code trajectories. These hints guide teacher-agent re-exploration, followed by rejection sampling to retain successful trajectories containing intermediate error feedback and recovery behaviors. (Bottom) Error-turn Masking Training removes explicit hints from student training inputs and masks the loss on erroneous action turns while retaining error feedback and subsequent recovery behaviors, enabling lightweight agents to learn error diagnosis and recovery without imitating faulty actions.}
\label{fig:sft}
\end{figure*}

As shown in Fig.~\ref{fig:sft}, the SFT phase in this paper designs a feedback-driven data construction and optimization framework centered on environmental feedback, comprising two key steps: Hint-guided Failure-aware Trajectory Generation and Error-turn Masking Training. Traditional agent SFT data construction typically relies on strong models to directly generate high-quality successful trajectories. However, for RS tool invocation tasks, due to the limited scale of available tasks, relying on strong models to generate complete multi-turn interaction trajectories at scale not only incurs high construction costs, but the generated expert trajectories also typically focus on ideal execution paths. This makes it difficult to cover anomalous feedback, state shifts, and error recovery processes in real tool environments, thereby limiting the generalization capabilities of lightweight agents in complex interaction scenarios.

To reduce the construction cost of high-quality training trajectories and fully utilize the feedback information generated in real tool environments, this paper proposes a feedback-driven SFT framework. This framework first extracts high-level instructional information---such as task constraints, tool dependencies, and error patterns---by analyzing the failure trajectories of the teacher agent during environmental interactions. It then utilizes the generated hint to guide the teacher agent in efficiently exploring and constructing training trajectories that contain environmental feedback. Subsequently, in the supervised learning phase, it further distinguishes between failed actions and valid recovery behaviors. It only masks the action supervision that leads to errors, while retaining the error feedback and the subsequent recovery process, enabling the model to learn how to recover from anomalous states while avoiding the imitation of invalid exploration behaviors.

\subsubsection{Hint-guided Failure-aware Trajectory Generation}

Specifically, for each training task, we first utilize a medium-scale open-source language model (the teacher agent) to conduct multi-turn autonomous interactions within the corresponding RS tool environment and collect its execution failure trajectories. Each training task independently undergoes the aforementioned interaction process to obtain task-specific failure states, tool invocation errors, and environmental feedback information.

Since the ground-truth trajectories provided by Earth-Bench primarily record explicit interface interaction relationships in the form of JSON tool invocation sequences, their ability to express data dependencies and continuous execution logic in complex tasks is relatively limited. Therefore, this paper further converts the official JSON ground truth into the corresponding ground-truth code trajectory to obtain a more compact and easily analyzable programmatic representation, which assists in the subsequent failure trajectory analysis and hint generation.

Subsequently, we input the failure trajectories collected for each task, the task descriptions, the corresponding ground-truth code trajectories, and their tool descriptions into the same open-source language model (hint generator), instructing it to generate an instructional hint specific to the current task under a hindsight analysis mode. The hint does not directly provide a complete executable solution or a specific sequence of domain-tool calls. Instead, it retains necessary control logic and task constraints, including task objectives, iteration strategies, required tool capabilities, data dependencies, intermediate state constraints, and potential execution risks. Concrete tool discovery and invocation remain the responsibility of the teacher agent.

Therefore, the role of the hint is not to replace the teacher agent's planning process, but to reduce the invalid exploration space and improve the sampling efficiency of successful trajectories while maintaining its autonomous tool discovery and environmental interaction capabilities.

This design has the following advantages:

\textbf{(1) Reducing training trajectory construction costs.} Compared to directly invoking high-capability models to generate complete expert trajectories, this paper solely utilizes open-source language models for failure interaction analysis and knowledge compression, and uses short-text hints to guide the teacher agent in autonomously sampling training trajectories, thereby lowering the trajectory construction costs.

\textbf{(2) Maintaining the autonomous interaction capability of the lightweight agent.} Although the ground-truth code trajectory provides the correct execution workflow, using it directly as the execution basis for the teacher agent would bypass the progressive tool discovery process in RS-Claw, causing the model to learn fixed execution paths rather than tool selection, state tracking, and environmental feedback response capabilities. By converting expert knowledge into abstract hints, this paper ensures that the teacher agent must still complete task planning and tool invocation based on environmental feedback.

\textbf{(3) Enhancing exploration efficiency using failure feedback.} Compared to extracting instructional information solely from successful trajectories, failure trajectories can provide information on tool misuse, parameter errors, and state shifts that occur during real interactions. By analyzing the differences between failure states and correct execution conditions, the hint can more specifically guide the teacher agent to avoid repeated invalid explorations.

After obtaining the hint, we input it as additional context into the initial state of the teacher agent, guiding it to re-explore the task execution process. For each task, we conduct multiple samplings and employ a rejection sampling strategy to retain the successful trajectories that ultimately meet the task requirements: $\hat{y}_{\tau} = y^\star$. Here, $y^\star$ represents the standard answer for the task, and $\hat{y}_{\tau}$ represents the final result produced by the trajectory.

It should be pointed out that the successful trajectories retained in this paper do not require all interaction turns to be optimal behaviors. Due to the dynamic feedback characteristics of real tool invocation environments, a successful trajectory may internally contain tool selection biases, code execution exceptions, or parameter adjustment processes. As long as the agent can complete state correction based on the error feedback returned by the environment and ultimately solve the task, the recovery processes within that trajectory are retained.

This design differs from supervision based solely on error-free expert trajectories. For lightweight agents, real deployment environments inevitably contain anomalous feedback and unknown states; merely learning ideal execution trajectories can easily lead to a lack of error diagnosis and recovery capabilities in the model. Therefore, this paper retains the complete interaction chain of "error occurrence---environmental feedback---state diagnosis---behavior correction," providing richer environmental interaction supervision signals for the subsequent Error-turn Masking training.

\subsubsection{Error-turn Masking Training}

To prevent the student model from directly fitting and imitating sub-optimal erroneous behaviors during the teacher's execution process in the SFT phase, we introduce the turn-level error masking mechanism shown in the lower part of Fig.~\ref{fig:sft} during the training phase. Its core idea is not to delete failed interactions, but to filter only the supervision signals corresponding to erroneous actions while retaining the complete environmental interaction history.

For time steps $t$ in a trajectory judged as erroneous exploration (i.e., satisfying $\mathcal{E}(a_t,o_t)=1$), we set the supervision mask corresponding to the action $a_t$ of that turn to 0, ensuring its generated tokens do not participate in the cross-entropy loss calculation:

\begin{equation}
L_{\text{mask}} = -\sum_{t=1}^{T} m_t \sum_{k=1}^{\vert{}a_t\vert{}} \log P_\theta(a_{t,k} \mid \tau_t, a_{t,<k})
\label{eq:2}
\end{equation}

where $k$ indexes tokens within action $a_t$, and $m_t$ is the turn-level supervision mask:

\begin{equation}
m_t = \begin{cases} 0, & \mathcal{E}(a_t,o_t)=1 \\ 1, & \mathcal{E}(a_t,o_t)=0 \end{cases}
\label{eq:3}
\end{equation}

This design ensures that erroneous actions $a_t$ do not contribute corresponding direct supervised losses, avoiding direct behavioral-cloning supervision on incorrect tool selection, invalid parameter configurations, or abnormal execution patterns.

It must be emphasized that Error-turn Masking only applies to the supervision targets and does not alter the trajectory context. For any time step $t$, the generation of the current action by the model is still based on the complete historical interaction:

\begin{equation}
P_\theta(a_t \mid \tau_t) = P_\theta(a_t \mid o_0, a_1, o_1, \dots, a_{t-1}, o_{t-1})
\label{eq:4}
\end{equation}

Therefore, after an erroneous action $a_t$ causes an environmental exception and generates feedback $o_t$, both the erroneous action and its corresponding environmental feedback are retained in the subsequent history to condition the learning of recovery behaviors:

\begin{equation}
P_\theta(a_{t+1} \mid \tau_{t+1}) = P_\theta(a_{t+1} \mid \tau_t, a_t, o_t)
\label{eq:5}
\end{equation}

where $o_t$ provides information such as environmental state shifts, tool execution results, and causes of exceptions, enabling the model to learn how to adjust its subsequent behaviors based on the feedback.

In other words, a training trajectory containing an error recovery process possesses an asymmetrical structure during supervised optimization:

\begin{itemize}
\item \textbf{Erroneous action }$a_t$: Retained in the context, but not used as a supervision target and contributes no corresponding direct supervised loss;
\item \textbf{Error feedback }$o_t$: Retained in the context, used to describe environmental state transitions;
\item \textbf{Recovery actions }$a_{t+1}, a_{t+2}, \dots$: Continue to serve as supervision targets to learn valid decision-making following errors.
\end{itemize}

Thus, this mechanism does not discard successful trajectories containing failed attempts and recovery processes, but decouples the different information components within failed interactions: the model avoids learning "how to generate errors" while simultaneously learning "how to understand errors and recover the task." Compared to training exclusively on idealized execution paths, this design leverages the "error occurrence---feedback observation---state correction---task recovery" process in real environmental interactions, improving the robustness of lightweight agents in complex tool environments.

Furthermore, when finally constructing the SFT data for the student model, we remove the explicit hint relied upon during the teacher model's sampling phase. This ensures the student model does not rely on extra cognitive scaffolding, but instead internalizes tool selection strategies, state tracking capabilities, and error recovery mechanisms from successful trajectories containing complete environmental feedback.

\subsection{Decision Evolution: Feedback-driven GRPO Optimization}

\begin{figure*}[!t]
\centering
\includegraphics[width=0.95\linewidth]{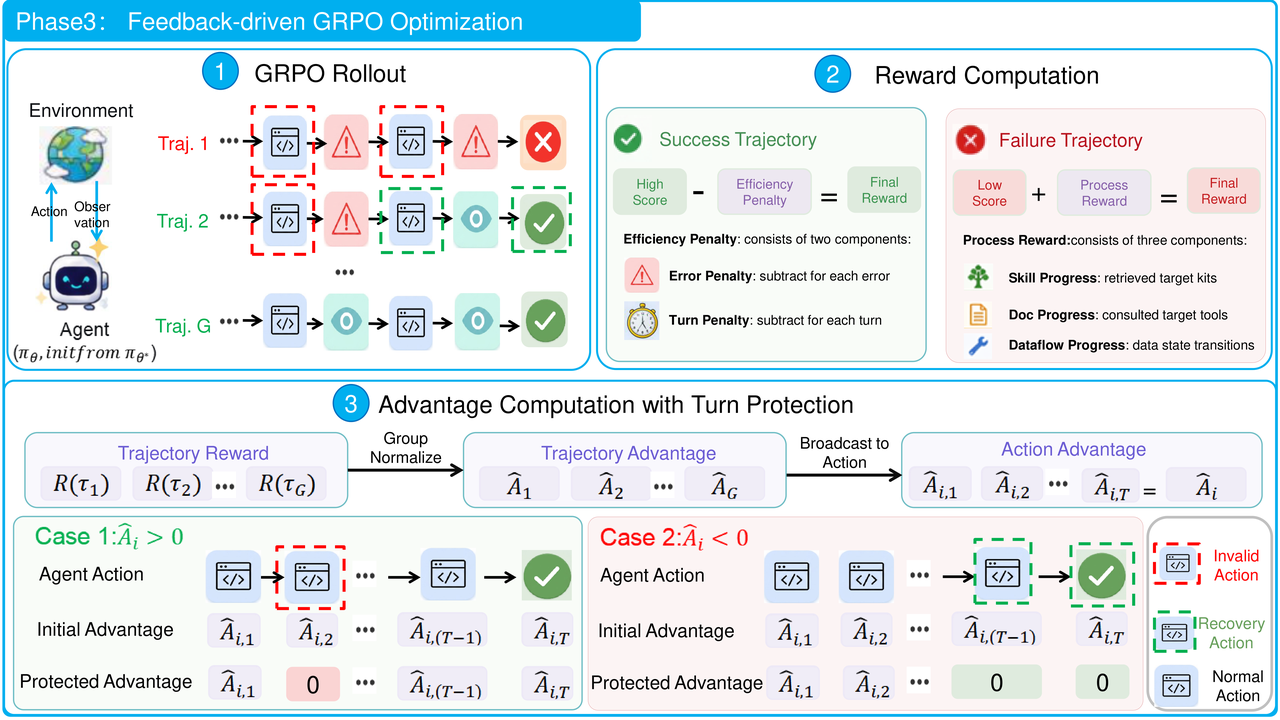}
\caption{Overview of the Decision Evolution stage in RS-Claw-Evolution: feedback-driven GRPO optimization for long-horizon RS agents. Compared with GRPO using sparse trajectory-level rewards, the proposed approach incorporates multi-dimensional process rewards derived from environmental feedback for finer-grained policy optimization. Furthermore, TAP selectively masks misleading turn-level advantages in successful trajectories containing erroneous exploration turns, reducing erroneous positive reinforcement while protecting recovery behaviors from inappropriate negative penalties. Together, these mechanisms support more reliable and efficient decision-making in long-horizon tool-use tasks.}
\label{fig:rl}
\end{figure*}

After endowing the model with basic tool invocation and error recovery capabilities through the SFT phase, this paper further employs RL to optimize the agent's long-term decision-making strategies in dynamic environments. As shown in Fig.~\ref{fig:rl}, the RL phase in this paper comprises three modules: GRPO trajectory sampling, MEFR calculation, and TAP.

This paper adopts GRPO~\cite{shao2024deepseekmath} as the foundational optimization algorithm. Unlike traditional value-network-based RL methods, GRPO estimates the advantage function through relative rewards among multiple trajectories sampled for the same task, thereby avoiding the overhead of training an extra value model. This is more suitable for optimizing long-horizon agent policies under limited interaction samples. It should be noted that this paper does not alter the basic optimization framework of GRPO, but redesigns the trajectory reward construction method and advantage propagation mechanism specifically for the reward sparsity and credit assignment problems present in RS tool invocation processes.

For a given RS task $q$, we use the old policy $\pi_{\theta_{\mathrm{old}}}$ to sample a group of $G$ complete interaction trajectories. Since rewards are evaluated at the sequence level, whereas the multi-turn interaction process consists of discrete decision steps, standard GRPO broadcasts the trajectory-level advantage derived from the global reward to local actions. Specifically, let $T_i$ denote the total number of interaction turns in the $i$th trajectory. Following the POMDP formalization in Section~\ref{sec:task-formalization}, at turn $t\in\{1,\ldots,T_i\}$, the model generates the complete action sequence $a_{i,t}$ conditioned on the historical interaction trajectory $\tau_{i,t}$. We initially assign the trajectory-level advantage to all turns by setting $\hat{A}_{i,t}=\hat{A}_i$ and optimize the policy using the clipped GRPO objective with Kullback--Leibler (KL) regularization:

\begin{subequations}\label{eq:6}
\begin{equation}
\mathcal{J}_{\mathrm{GRPO}}(\theta)=\mathbb{E}\!\left[\frac{1}{G}\sum_{i=1}^{G}\frac{1}{T_i}\sum_{t=1}^{T_i}\left(\mathcal{L}_{i,t}^{\mathrm{clip}}-\beta\operatorname{KL}_{i,t}\right)\right]
\label{eq:6a}
\end{equation}
\begin{equation}
\mathcal{L}_{i,t}^{\mathrm{clip}}=\min\!\left(\rho_{i,t}\hat{A}_{i,t},\operatorname{clip}\!\left(\rho_{i,t},1-\epsilon,1+\epsilon\right)\hat{A}_{i,t}\right)
\label{eq:6b}
\end{equation}
\end{subequations}

Here, $\rho_{i,t}=\pi_\theta(a_{i,t}\mid\tau_{i,t})/\pi_{\theta_{\mathrm{old}}}(a_{i,t}\mid\tau_{i,t})$ is the probability ratio for generating action $a_{i,t}$ at turn $t$, and $\epsilon$ is the clipping hyperparameter used to limit the magnitude of policy updates. The turn-level penalty $\operatorname{KL}_{i,t}=D_{\mathrm{KL}}\!\left(\pi_\theta(\cdot\mid\tau_{i,t})\,\|\,\pi_{\mathrm{ref}}(\cdot\mid\tau_{i,t})\right)$ constrains the learned policy relative to the frozen reference policy $\pi_{\mathrm{ref}}$, which is initialized from the same SFT checkpoint, while $\beta$ controls the strength of this regularization.

The sequence-level advantage estimate $\hat{A}_i$ corresponding to the $i$th trajectory is calculated by normalizing the total rewards of the $G$ trajectories within the same group:

\begin{equation}
\hat{A}_i = \frac{R(\tau_i) - \text{mean}(R(\tau_1), \ldots, R(\tau_G))}{\text{std}(R(\tau_1), \ldots, R(\tau_G)) + \delta}
\label{eq:7}
\end{equation}

where $\delta>0$ is a small constant introduced for numerical stability when the within-group reward standard deviation is zero or close to zero.

However, for long-horizon RS agents, indiscriminately broadcasting this trajectory-level advantage to all action turns, i.e., assigning $\hat{A}_{i,t}=\hat{A}_i$, has two limitations:

\begin{enumerate}
\item Final outcome-level rewards fail to reflect local contributions during complex tool invocation processes. A failed trajectory might have completed correct tool retrieval, document understanding, or partial data processing, but due to a final incorrect answer, it receives the same low reward as invalid exploration.
\item Trajectory-level advantage propagation ignores state dependencies between actions. When a trajectory contains erroneous exploration and recovery behaviors, uniform advantage propagation may lead to valid recovery actions being penalized by errors, or erroneous exploration behaviors being reinforced by advantages.
\end{enumerate}

Therefore, this paper extends GRPO from two perspectives: reward modeling and advantage propagation.

\subsubsection{MEFR}

In the RL phase, to avoid the costs and evaluation biases introduced by training extra evaluation models, this paper does not introduce an independent process reward model. Instead, it directly utilizes the structured feedback generated by the RS execution environment to construct process-level rewards. Specifically, tool catalog matching, document understanding, and data flow execution states are all determined by environmental state transitions, rather than relying on model generation results for evaluation.

As shown in Fig.~\ref{fig:rl} (top right), we discard the single binary outcome reward and construct an MEFR function capable of finely distinguishing trajectory quality. For trajectories that ultimately fail to complete the task, on top of the base failure score, we introduce dense rewards for skill catalog retrieval, tool document reading, and data flow progress. This quantifies the proportion of sub-tasks the model has correctly completed, providing more discriminative rewards for partially correct explorations rather than assigning all failed trajectories the same score. For successful trajectories that have reached the correct solution, on top of the base success score, we introduce absolute turn penalties and error penalties to suppress redundant invocations and explicit mistakes during the process, encouraging the model to explore more efficient successful execution strategies.

The reward function is defined as:

\begingroup
\fontsize{7.25pt}{8.75pt}\selectfont
\begin{equation}
R(\tau) = \begin{cases}
\alpha_{\text{success}} - \lambda T(\tau) - \mu E(\tau), & \text{if } \hat{y}_{\tau} = y^\star, \\
\alpha_{\text{fail}} + \omega_{\text{skill}} r_{\text{skill}}(\tau) + \omega_{\text{doc}} r_{\text{doc}}(\tau) + \omega_{\text{data}} r_{\text{data}}(\tau), & \text{if } \hat{y}_{\tau} \neq y^\star.
\end{cases}
\label{eq:8}
\end{equation}
\endgroup

Here, $\alpha_{\text{success}}$ and $\alpha_{\text{fail}}$ represent the base score parameters for task success and failure, respectively; $\lambda, \mu \ge 0$ are the weights for the constant deduction terms; $\omega_{\text{skill}}, \omega_{\text{doc}}, \omega_{\text{data}} \ge 0$ are the weights for the progress rewards. This formulation reflects a clear reward shaping mechanism: when a trajectory fails, the model starts from $\alpha_{\text{fail}}$ and continuously accumulates positive scores by retrieving correct tool catalogs, reading corresponding documents, and executing correct data flows; when a trajectory succeeds, the model receives a high $\alpha_{\text{success}}$, and redundant turns and errors result in direct score deductions.

The specific calculation methods for each dense progress sub-term are as follows:

\begingroup
\fontsize{8pt}{9.5pt}\selectfont
\begin{equation}
r_{\text{skill}}(\tau)=\frac{\vert{}\mathcal{S}_{\tau}\cap\mathcal{S}\vert{}}{\vert{}\mathcal{S}\vert{}},\;
r_{\text{doc}}(\tau)=\frac{\vert{}\mathcal{D}_{\tau}\cap\mathcal{D}\vert{}}{\vert{}\mathcal{D}\vert{}},\;
r_{\text{data}}(\tau)=\frac{\vert{}\mathcal{G}_{\tau}\cap\mathcal{G}\vert{}}{\vert{}\mathcal{G}\vert{}}.
\label{eq:9}
\end{equation}
\endgroup

Since all tasks have ground-truth solutions and must go through multi-step tool invocations to be completed, the target set cardinalities in the denominators, $\vert{}\mathcal{S}\vert{}$, $\vert{}\mathcal{D}\vert{}$, and $\vert{}\mathcal{G}\vert{}$, are all strictly greater than 0.

$\mathcal{S}$, $\mathcal{D}$, and $\mathcal{G}$ represent the target tool catalog (Skill) set, target tool document (Doc) set, and target data flow steps set in the reference workflow for solving the problem, respectively. The subscripted $\mathcal{S}_{\tau}$, $\mathcal{D}_{\tau}$, and $\mathcal{G}_{\tau}$ correspond to the respective sets actually triggered by the agent in the current trajectory $\tau$ (detailed determination methods are provided in Appendix D of the Supplementary Material). By calculating the intersection, we effectively filter out hallucinations and redundant invocations generated by the agent during exploration, ensuring the model only gains progress scores when hitting key nodes in the reference workflow.

Meanwhile, $T(\tau)$ represents the total number of assistant turns in the trajectory, and $E(\tau)$ represents the absolute count of attributable errors such as explicit invalid actions, code execution errors, format errors, and repetitive invocations. Due to the constant deduction design applied to successful trajectories, the model faces clear constraints, thereby spontaneously reducing $T(\tau)$ and $E(\tau)$ while ensuring correctness.

To prevent the model from acquiring false rewards through invalid exploration, all our environment feedback rewards are calculated based on environmental state transitions rather than language outputs. Document and data flow progress calculations are evaluated entirely based on state transitions in the runtime environment. For document and skill progress, we take the pre-set target tool set in the reference workflow as the prior state. During the trajectory evaluation phase, the interactive environment automatically parses the atomic action instructions generated by the agent across reasoning turns (including tool catalog retrieval, instruction document reading, and actual tool invocation). Only when the underlying skill domain probed or the tool document read by the agent strictly belongs to the target reference set will the corresponding action be updated into the actually triggered sets $\mathcal{S}_{\tau}$ and $\mathcal{D}_{\tau}$. For data flow progress, we introduce a fine-grained state verification mechanism based on data dependency graphs. For complex multi-step RS tasks, we pre-extract benchmark data flow steps containing state transition constraints based on standard solutions pre-set in the evaluation benchmark. Each node in this benchmark not only defines the target tool to be invoked but also strictly constrains the expected input/output data family features and dependencies of intermediate products. During training, the environment dynamically tracks and aggregates the actual consumed input states and generated output states within the agent's interaction trajectory. Only when a tool's runtime record fully covers the data constraints of a benchmark node and successfully maintains the necessary upstream data logic will that step be deemed validly completed and counted into $\mathcal{G}_{\tau}$.

\subsubsection{TAP}

In standard GRPO, the trajectory-level relative advantage $\hat{A}_i$ is assigned uniformly to all turns, such that $\hat{A}_{i,t}=\hat{A}_i$ in~\eqref{eq:6b}. However, for RS tasks involving multi-turn tool interactions, the contribution of each turn to the final result varies significantly. This uniform assignment introduces two credit-assignment problems:

\begin{itemize}
\item \textbf{Error-Recovery Penalty:} When a successful trajectory is judged as a within-group disadvantage ($\hat{A}_i < 0$) due to exploration redundancy, it means it underwent extensive "trial-and-error $\rightarrow$ feedback $\rightarrow$ correction" processes. If the global negative advantage is directly broadcast to those "correction turns" that pulled the task back on track, the model receives an erroneous penalty signal, thereby learning that "one should not continue correcting after an error." This makes the model highly susceptible to stalling or giving up prematurely when encountering environmental anomalies in the future.
\item \textbf{Trial-and-Error Solidification:} When a successful trajectory is judged as a relative advantage ($\hat{A}_i > 0$), it may still internally contain localized erroneous turns. In complex tasks, these errors are often necessary explorations for the model to actively probe unknown tool constraints. If these actions are allowed to inherit a global positive advantage, the model will develop a false alignment of "deliberately outputting erroneous actions."
\end{itemize}

To mitigate these issues, we introduce the TAP mechanism, as shown in Fig.~\ref{fig:rl} (bottom). Its core idea is: for trajectories that ultimately yield correct results, whether it is localized erroneous exploration in a positive-advantage trajectory or successful error-correction turns in a negative-advantage trajectory, their corresponding local advantages are forcefully truncated and masked to 0, thereby blocking the propagation of misleading gradients. The mathematical expression of this mechanism is as follows:

\begingroup
\footnotesize
\begin{equation}
A_{i,t} = \begin{cases}
0, & \hat{y}_{\tau_i} = y^\star \land \left[(\hat{A}_i > 0 \land E_{i,t}) \lor (\hat{A}_i < 0 \land R_{i,t})\right] \\
\hat{A}_i, & \text{otherwise}
\end{cases}
\label{eq:10}
\end{equation}
\endgroup

The relevant indicator functions are defined as follows:

\begin{itemize}
\item $E_{i,t} = \mathcal{E}(a_{i,t}, o_{i,t})$ is the global error indicator function defined in Section~\ref{sec:task-formalization}.
\item $R_{i,t} = \text{IsRecovery}(t)$ is the error-correction turn indicator function. For an ultimately successful trajectory $\tau_i$, we define the subsequent interaction turns that follow the last error feedback and ultimately contribute to task completion as the recovery sequence.
\end{itemize}

This mechanism does not re-estimate the group-relative advantage. Instead, $A_{i,t}$ from~\eqref{eq:10} replaces $\hat{A}_{i,t}$ in the clipped objective~\eqref{eq:6b}. For ultimately successful trajectories, this suppresses negative policy-surrogate updates on identified recovery turns when $\hat{A}_i<0$, and positive policy-surrogate updates on erroneous turns when $\hat{A}_i>0$. All optimized tokens within the same assistant turn share the corresponding protected turn-level advantage $A_{i,t}$; the KL regularization term is unchanged.

\section{Experiments}

\subsection{Experimental Setup}

\textbf{Dataset and Evaluation Scope.} We conduct our experimental evaluation on the RS agent benchmark Earth-Bench~\cite{feng2026earthagent}. This benchmark contains 248 RS analysis tasks covering workflows across various RS domains such as atmospheric RS, urban and human activities, and land surface thermal environments. It provides two evaluation modes: Autonomous Planning (AP) and Instruction Following (IF). The AP mode requires the agent to autonomously plan the tool invocation workflow based solely on the task objective, primarily evaluating capabilities in task decomposition, tool discovery, and execution decision-making. The IF mode provides explicit execution instructions, mainly testing the agent's ability to translate given steps into executable tool invocations.

Since this paper focuses on planning, state tracking, and environmental feedback learning capabilities during long-horizon multi-tool interactions, we select a subset of 188 tasks from Earth-Bench that contain real tool execution workflows for our experiments. We further select 12 real RS tasks from this subset for training data construction, completely holding out the remaining 176 tasks for testing. During testing, the full tool library setup is maintained; the agent still faces the original tool scale, tool discovery process, and tool document organization. A detailed analysis of the test set filtering criteria and excluded tasks is provided in Appendix A of the Supplementary Material.

\textbf{Training Task Coverage.} According to the reference execution workflows provided by Earth-Bench, the selected training tasks involve 28 different tools, accounting for approximately 27\% of the complete tool library. It should be noted that this statistic only reflects the tool coverage in the reference trajectories and does not imply that the agent is restricted to accessing only these tools during the training phase. Since the agent still conducts interactive trajectory sampling within the full tool library environment, it may explore tool combinations during training that do not appear in the reference workflows.

The held-out test set is used to evaluate the model's generalization capabilities to unseen tasks, novel tool combinations, and long-horizon workflows under limited task demonstration conditions. Detailed information regarding the training task selection strategy and task distribution can be found in Appendix A of the Supplementary Material.

\textbf{Models and Interaction Paradigms.} To verify the effectiveness of the proposed method on lightweight language models, we primarily select Qwen3-4B-Instruct and Qwen3-8B as base models, evaluating their original models, SFT models, and RL-optimized models under the RS-Claw~\cite{liu2026rsclaw} Code interaction paradigm. Additionally, to measure the performance gap between lightweight models and large-scale models, we introduce Qwen3-32B, DeepSeek-V3.1, and GPT-5 as strong model references. Since the original RS-Claw experiments were mainly based on a structured JSON tool invocation paradigm, we additionally report JSON interaction results as a paradigm baseline to analyze the impact of Code-based action representations on long-horizon tool invocation. These results are re-aggregated from the official per-task results by selecting the same 176 held-out test tasks used in our evaluation. In this paper, the \emph{Code} paradigm indicates that the model uses continuous Python code for tool exploration and invocation; the \emph{JSON} paradigm denotes the structured tool invocation format used in the original RS-Claw results.

\textbf{Training Parameters.} Both the SFT and RL stages are implemented on top of the slime framework~\cite{zhu2025slime}. All experiments are conducted on NVIDIA A100 graphics processing units (GPUs).

\emph{SFT phase.} We train both models for 2 epochs using a global batch size of 4. The learning rate is initialized to $5 \times 10^{-6}$ and follows a cosine decay schedule down to a minimum of $5 \times 10^{-7}$.

\emph{RL phase.} For each prompt, we sample a group of 12 rollouts (group size $G = 12$) to estimate the advantage. The reward function parameters are empirically set to balance task completion and efficiency: the base success reward $\alpha_{\text{success}} = 10$, base failure penalty $\alpha_{\text{fail}} = -1.0$, turn penalty $\lambda = 0.1$, and error penalty $\mu = 0.25$. For the MEFR, the weights are set to $\omega_{\text{data}} = 1.0$ for dataflow progress, $\omega_{\text{doc}} = 0.25$ for document reading progress, and $\omega_{\text{skill}} = 0.2$ for skill directory progress. The RL training runs for 5 epochs with a constant learning rate of $1 \times 10^{-6}$. Detailed training parameter settings and justifications are provided in Appendix B of the Supplementary Material.

\textbf{Evaluation Parameters.} All evaluation results are obtained using a temperature of 0.1, with other parameters kept at their default values. The model's action space is capped at 25 execution steps, meaning a maximum of 25 interaction turns are allowed for a single task.

\textbf{Evaluation Metrics.} We adopt the end-to-end metrics and process-level tool invocation metrics provided by Earth-Bench for evaluation. End-to-end metrics include \emph{Accuracy} and \emph{Efficiency}, where Accuracy measures the consistency between the agent's final answer and the standard answer, and Efficiency measures the execution efficiency of the agent's actual number of tool invocations relative to the reference trajectory. Process-level metrics include \emph{Tool-Any-Order}, \emph{Tool-In-Order}, \emph{Tool-Exact-Match}, and \emph{Parameters} (parameter accuracy), which are used to assess the agent's capabilities in tool selection, execution sequence, and parameter generation.

Furthermore, to analyze the impact of different interaction paradigms (Code vs. JSON) on the execution overhead of long-horizon RS tasks, we track the average number of generated \emph{Action Tokens} and environment-returned \emph{Observation Tokens} for each task. This measures the compression effect of Code-based interaction on action representations and environmental feedback contexts.

Focusing on execution reliability and interaction efficiency in multi-turn tool invocations, this paper further introduces two metrics: \emph{Error Interaction Rate (EIR)} and \emph{Avg Turns}. EIR measures the proportion of actions that trigger environmental error feedback during the agent's interaction process, defined as the ratio of the number of turns where $\mathcal{E}(a_t, o_t) = 1$ to the total number of assistant action turns. Avg Turns represents the average number of interaction turns generated by the assistant in a single task, used to measure the agent's execution efficiency in completing the task. Full metric definitions are provided in Appendix A, while the complete results are reported in Appendix G of the Supplementary Material.

\subsection{Overall Performance Comparison}

\begin{table*}[!t]
\centering
\caption{Overall Performance Comparison of Language Models with Varying Scales on Earth-Bench. "Code" indicates the programmatic interaction paradigm utilized in our method, while "JSON" represents the traditional structured tool-calling paradigm used in the original RS-Claw framework.}
\label{tab:table1}
\footnotesize
\setlength{\tabcolsep}{4pt}
\renewcommand{\arraystretch}{1.08}
\begin{tabular}{llcccccc}
\toprule
\multirow{2}{*}{\textbf{Model}} & \multirow{2}{*}{\textbf{Paradigm}} & \multicolumn{2}{c}{\textbf{Accuracy}} & \multicolumn{2}{c}{\textbf{Tool-Any-Order}} & \multicolumn{2}{c}{\textbf{Tool-In-Order}} \\
\cmidrule(lr){3-4} \cmidrule(lr){5-6} \cmidrule(lr){7-8}
& & \textbf{AP} & \textbf{IF} & \textbf{AP} & \textbf{IF} & \textbf{AP} & \textbf{IF} \\
\midrule
Qwen3-4B-Instruct & Code & 31.8 & 35.8 & 47.6 & 52.8 & 40.1 & 45.8 \\
Qwen3-4B SFT & Code & 41.5 & 44.3 & 58.6 & 60.9 & 49.4 & 51.8 \\
Qwen3-4B RL (Ours) & Code & \textbf{65.9} & \textbf{65.9} & \textbf{69.1} & 70.1 & \textbf{58.9} & 58.4 \\
Qwen3-8B & Code & 19.3 & 22.7 & 49.6 & 50.9 & 41.1 & 44.2 \\
Qwen3-8B SFT & Code & 46.6 & 44.3 & 61.4 & 63.6 & 52.2 & 53.2 \\
Qwen3-8B RL (Ours) & Code & 61.4 & 64.2 & 67.8 & \textbf{70.5} & 57.8 & \textbf{60.0} \\
Qwen3-32B & Code & 43.8 & 42.6 & 59.3 & 60.3 & 47.8 & 50.7 \\
DeepSeek-V3.1 & Code & 60.8 & 65.3 & \textbf{71.2} & \textbf{73.1} & \textbf{60.0} & \textbf{62.4} \\
GPT-5 & Code & \textbf{71.6} & \textbf{71.0} & 69.1 & 72.0 & 58.4 & 61.8 \\
Qwen3-32B & JSON & 35.8 & 34.1 & 54.1 & 63.4 & 36.7 & 50.0 \\
DeepSeek-V3.1 & JSON & 55.7 & 55.1 & \textbf{78.7} & \textbf{79.6} & \textbf{65.0} & \textbf{66.1} \\
GPT-5 & JSON & \textbf{68.8} & \textbf{71.6} & 72.4 & 75.2 & 58.5 & 61.0 \\
\bottomrule
\end{tabular}
\end{table*}

Table~\ref{tab:table1} summarizes the overall execution performance of language models of varying scales on Earth-Bench. Overall, RS-Claw-Evolution significantly enhances the execution capabilities of lightweight language models in complex RS tool invocation tasks. This indicates that by utilizing feedback signals generated during environmental interactions for training, the model's task planning and execution capabilities in long-horizon, multi-tool environments can be effectively improved.

First, the SFT phase can significantly improve the RS tool-use capabilities of the base models. Taking Qwen3-4B-Instruct as an example, after SFT, the Accuracy in AP mode increases from 31.8\% to 41.5\%, Tool-Any-Order increases from 47.6\% to 58.6\%, and Tool-In-Order increases from 40.1\% to 49.4\%. Similar improvements are also observed in the Qwen3-8B model, indicating that even by utilizing only a small number of real RS tasks, the trajectory construction strategy guided by failure feedback can still effectively enhance the model's capability to model tool semantics, task workflows, and state transition regularities.

Second, the RL phase further enhances the model's decision-making capabilities in open-ended AP scenarios. For Qwen3-4B-Instruct, the Accuracy in AP mode reaches 65.9\% after RL, an improvement of 24.4 percentage points compared to the SFT phase; in IF mode, it also reaches 65.9\%, an improvement of 21.6 percentage points compared to the SFT phase. Simultaneously, the Tool-In-Order metric increases from 49.4\% to 58.9\%, indicating that the process-level optimization signals constructed based on environmental feedback not only improve the final task completion rate but also further optimize the tool execution order and strategy selection in multi-step RS workflows. These results demonstrate that through process-level rewards and the TAP mechanism, the model can reduce invalid explorations and gradually learn more stable long-horizon interaction strategies.

Third, after feedback-driven training, lightweight models can approach or even surpass the execution level of some larger-scale language models. Taking Qwen3-4B RL as an example, its AP Accuracy reaches 65.9\%, significantly outperforming the untrained Qwen3-32B (43.8\%), and reaching or exceeding the execution level of some large-scale general-purpose models---for instance, surpassing DeepSeek-V3.1 under the Code paradigm (60.8\%), and only slightly lower than GPT-5 (71.6\%). Meanwhile, Qwen3-8B RL also achieves an AP Accuracy of 61.4\%, a further improvement of 14.8 percentage points compared to its SFT phase. This demonstrates that feedback-driven training is not only applicable to small-scale models but can stably enhance the environmental interaction and autonomous decision-making capabilities of models across different parameter scales. These results suggest that in real RS tool invocation scenarios, model scale is not the sole determinant of performance; effectively optimizing the environmental interaction process can significantly compensate for the deficiencies of lightweight models in tool understanding, state tracking, and error recovery.

Notably, after RL optimization, the final performance of Qwen3-4B is slightly higher than that of Qwen3-8B. We attribute this phenomenon primarily to the RL phase's dependence on initial policy quality and optimization stability. Compared to simply increasing model scale, performance gains in the RL phase depend more on the model's utilization efficiency of environmental feedback and the optimization space of its existing tool invocation strategies. Due to differences in tool-use priors formed during the pre-training and instruction fine-tuning phases across models of different scales, a larger parameter scale does not necessarily directly translate to higher environmental interaction benefits.

Further comparison of different tool interaction paradigms reveals that programmatic Code interaction yields varying degrees of gains for language models at different capability levels. Its advantage is mainly reflected in alleviating the state management burden of lightweight models in long-horizon, multi-tool environments. Compared to the RS-Claw JSON paradigm, the AP Accuracy of Qwen3-32B under the Code paradigm increases from 35.8\% to 43.8\%, DeepSeek-V3.1 from 55.7\% to 60.8\%, while GPT-5 only increases from 68.8\% to 71.6\%. This trend indicates that Code interaction does not simply enhance the task capabilities of all models, but primarily mitigates the limitations brought by explicit state representation and context accumulation during the tool invocation process.

For highly capable LLMs, they inherently possess strong long-context understanding and complex state tracking capabilities. Therefore, redundant tool descriptions, file paths, and intermediate state information in JSON tool invocations have a relatively limited impact on their final decisions, resulting in a corresponding decrease in the additional benefits brought by the Code paradigm. However, for parameter-constrained models, a large amount of explicit tool information continuously occupies the limited context space and increases the difficulty of state maintenance in multi-turn interactions. By managing intermediate states through program variables and organizing continuous operations using code logic, the Code paradigm can lower state maintenance costs, enabling lightweight models to allocate more context capacity to task planning, tool selection, and error recovery. Consequently, the combination of programmatic interaction and environmental feedback learning can significantly narrow the capability gap between lightweight models and large-scale models in complex RS tasks.

Further analysis of the tool execution metrics reveals that the Tool-Any-Order metric for some models under the Code paradigm is slightly lower than under the JSON paradigm. This does not imply that programmatic interaction reduces tool-use capability; rather, it primarily stems from differences in operational granularity and state representation between the two paradigms. The Earth-Bench tool library contains some basic calculation and data processing tools, such as numerical operations and statistical aggregations. Under the JSON paradigm, the agent needs to explicitly invoke these tools, so the relevant operations appear in the predicted tool sequence and are included in the tool matching metric calculation. In contrast, under the Code paradigm, such basic operations can typically be completed directly via Python expressions or intermediate variables without generating corresponding explicit tool invocations, and thus are not factored into the Tool-Any-Order matching results.

In comparison, the Tool-In-Order metric remains generally stable under the Code paradigm and even shows improvement on some models. This indicates that programmatic interaction does not disrupt the critical RS tool invocation workflow, but instead reduces the explicit state transmission burden through more flexible programmatic state management. This result also demonstrates that for long-horizon RS tasks, strict matching with reference tool sequences is not the sole standard for measuring an agent's capability; the ability to select reasonable execution paths based on task states and achieve the objective is equally a key factor in measuring agent autonomy.

\subsection{Context Efficiency Analysis}

To evaluate whether programming-based interaction alleviates context overhead in long-horizon RS tasks, we compare runtime interaction costs between Code and JSON paradigms, including action representation overhead (Action Tokens) and environmental feedback overhead (Observation Tokens).

Note that Action Tokens here measure runtime tool interaction costs rather than SFT training tokens. Specifically, for Code, Action Tokens are computed from executable Python code blocks generated by the agent, while for JSON, they are computed from structured tool names and parameters. Observation Tokens measure the actual environmental feedback returned to the model after tool execution, including tool outputs, program outputs, and execution exceptions. All statistics are calculated using a unified tokenizer while excluding system prompts, task descriptions, reasoning traces, and final answers. Token counts are accumulated over each complete task trajectory and then averaged across evaluation tasks; structured tool calls are serialized into compact JSON before tokenization.

\begin{figure}[!t]
\centering
\includegraphics[width=\linewidth]{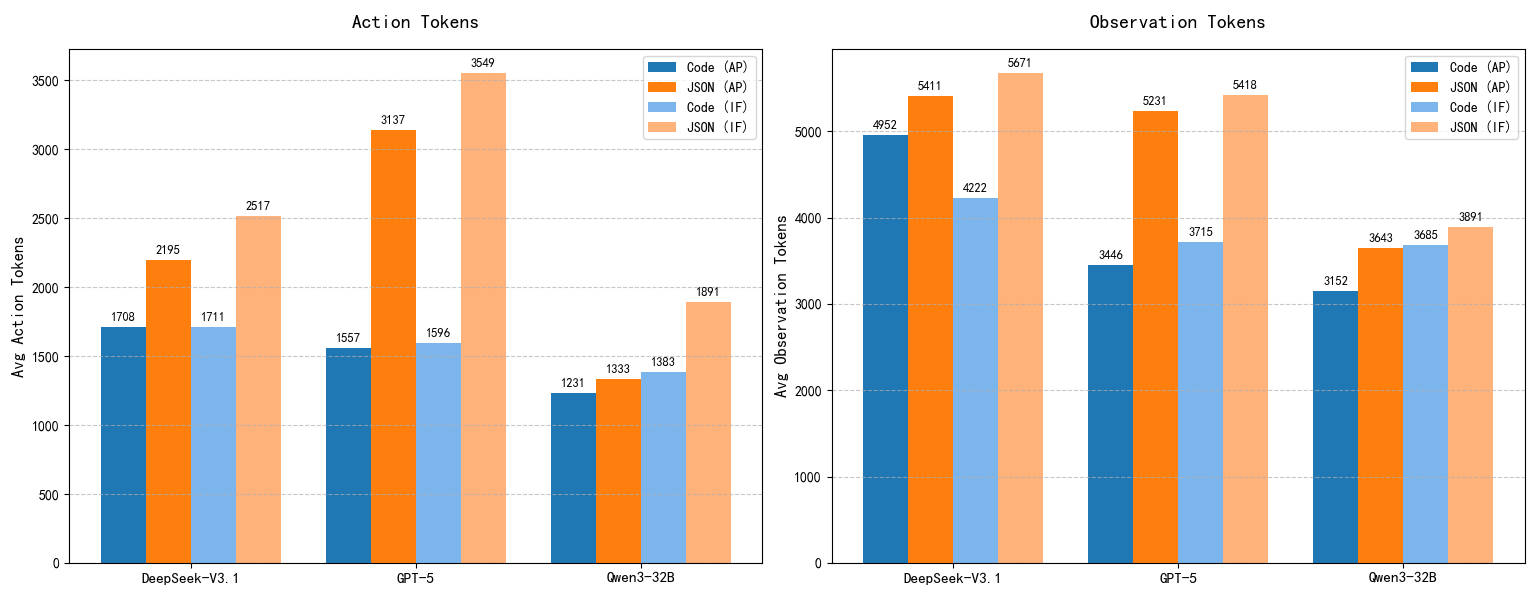}
\caption{Comparison of context efficiency between Code and JSON interaction paradigms. The left and right panels respectively show the average Action Tokens and Observation Tokens consumed in a single trajectory by different language models (DeepSeek-V3.1, GPT-5, and Qwen3-32B). The results indicate that compared to the traditional structured JSON tool invocation paradigm, the programmatic interaction paradigm (Code) adopted in this paper effectively reduces context overhead in both AP and IF modes.}
\label{fig:context}
\end{figure}

Fig.~\ref{fig:context} illustrates the average Action Tokens and Observation Tokens under different interaction paradigms on Earth-Bench. The results demonstrate that the Code paradigm consistently reduces interaction overhead across the evaluated models in both AP and IF modes.

First, Code substantially reduces Action Tokens by replacing repeated tool specifications and parameter transmission with programmatic state management. In JSON-based interaction, the agent must explicitly serialize tool calls and associated states at every turn, whereas Code enables intermediate states to be stored and reused through variables and control flows.

Further comparison reveals that the action-token savings from Code interaction vary substantially across models. In AP mode, GPT-5's Action Tokens decrease from 3137.1 under JSON interaction to 1557.2 under Code interaction, a reduction of approximately 50.4\%. By comparison, Qwen3-32B's Action Tokens decrease from 1333.0 to 1231.0, corresponding to a reduction of only 7.7\%. One possible explanation is that models differ in how effectively they exploit variable references, control flows, and intermediate-state reuse to produce compact action sequences. However, these aggregate statistics do not isolate programmatic planning capability from differences in trajectory length, tool selection, or execution outcomes. The results therefore suggest that the compression benefits of Code interaction depend partly on how the model uses the representation, rather than on the action format alone.

Second, Code also reduces Observation Tokens by changing how intermediate states are exposed to the agent. Unlike JSON interaction, where tool outputs are directly returned after each invocation, Code allows temporary intermediate results to remain inside the execution environment and only exposes necessary states, explicit outputs, and exceptions to the model.

For instance, in AP mode, Observation Tokens decrease from 5230.8 to 3446.4 for GPT-5 and from 3642.7 to 3151.7 for Qwen3-32B. These results indicate that programming-based interaction not only compresses action representations but also enables more selective environmental feedback transmission.

Overall, the benefits of Code interaction arise from two complementary mechanisms: compact action abstraction and implicit state maintenance through program execution. These mechanisms motivate the use of programming-based interaction to reduce the context burden of lightweight agents in long-horizon RS workflows.

\subsection{Ablation Study on Failure-aware Supervision}

All ablation experiments are evaluated in AP mode.

\subsubsection{Impact of Hindsight Hint-guided Trajectory Generation}

To verify whether the proposed Hint-guided Failure-aware Trajectory Generation method can improve the construction efficiency of training trajectories under a limited interaction budget, we conducted controlled experiments on the 12 training tasks.

Specifically, we employ Qwen3-32B as both the teacher agent and the hint generator to generate training trajectories under the same tool environment and sampling configurations. For each training task, we compare two trajectory generation strategies:

\begin{itemize}
\item \textbf{Without Hint:} Only the original task description and tool environment are provided, and the teacher agent autonomously explores the task execution workflow.
\item \textbf{With Hindsight Hint:} A hindsight hint, derived from the analysis of failed execution trajectories, is added to the initial context. This hint does not directly provide a complete executable solution or a specific sequence of domain-tool calls; instead, it retains necessary control logic and task constraints, including task objectives, required tool capabilities, data dependencies, and potential execution risks. It is used to guide the teacher agent to avoid repeated invalid explorations and improve the probability of generating successful trajectories.
\end{itemize}

For both settings, 100 trajectories are independently sampled for each training task, and the proportion of trajectories that successfully complete the task is calculated to measure the efficiency of acquiring valid training data under a fixed sampling budget across different trajectory generation strategies.

\begin{figure}[!t]
\centering
\includegraphics[width=\linewidth]{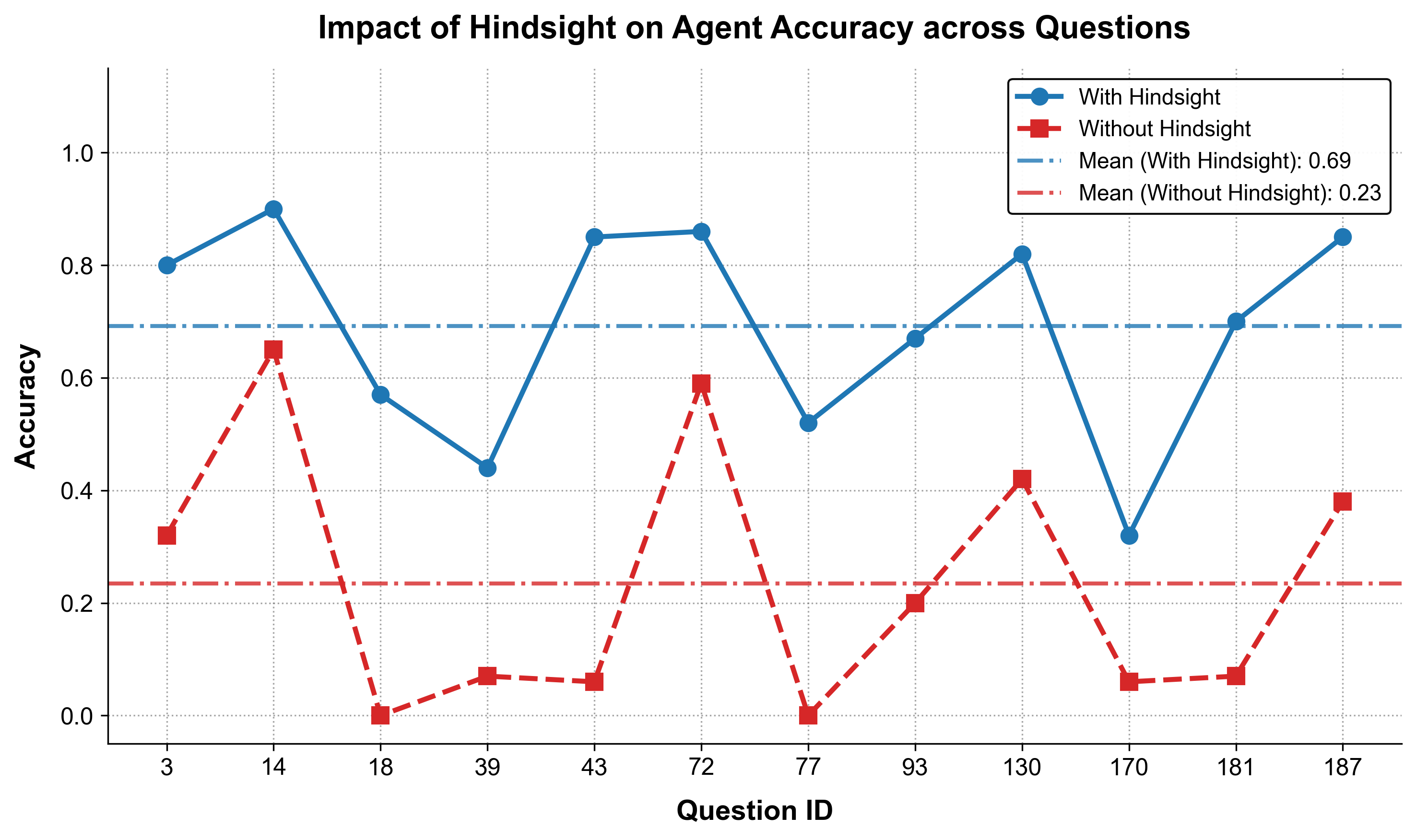}
\caption{Impact of the Hindsight Hint guidance mechanism on the trajectory generation success rate of 12 training tasks. The solid blue line and dashed red line represent the exploration success rates of the teacher agent with and without the Hindsight Hint guidance, respectively.}
\label{fig:hint}
\end{figure}

As shown in Fig.~\ref{fig:hint}, after introducing the Hindsight Hint, all 12 training tasks exhibit varying degrees of improvement in sampling success rates. The overall average success rate increased from 23.5\% to 69.3\%, indicating that instructional information generated using failed execution experiences can effectively narrow the invalid exploration space of the teacher agent and increase the collection probability of valid training trajectories in complex RS tasks.

Further analysis of different tasks reveals that the benefits of the Hindsight Hint are more pronounced in complex long-horizon tasks. For example, Tasks 18 and 77 failed to yield any valid successful trajectories under the \emph{Without Hint} condition, but their success rates increased to 57\% and 52\%, respectively, after incorporating the Hindsight Hint. For tasks with initially low success rates, the success rates of Tasks 39 and 43 also increased from 7\% and 6\% to 44\% and 85\%, respectively. These tasks typically involve multi-stage data processing, complex tool combinations, and strict data flow dependencies. When relying solely on autonomous exploration, the teacher agent is prone to entering inefficient execution paths due to early tool selection errors, parameter configuration deviations, or insufficient state understanding. Furthermore, even for tasks where the agent already possesses a certain level of execution capability, the Hindsight Hint can still further reduce invalid explorations. For instance, the success rates of Tasks 14 and 72 increased from 65\% and 59\% to 90\% and 86\%, respectively.

These results demonstrate that the Hindsight Hint does not simply provide a fixed execution scheme; rather, by analyzing error patterns in failed interactions and integrating the task dependency constraints provided by the ground-truth code trajectory, it compresses specific interaction experiences into high-level exploration guidance. While maintaining the teacher agent's capability for autonomous tool discovery and environmental feedback response, the hint helps the agent avoid repeated invalid attempts, thereby enhancing the acquisition efficiency of successful training trajectories under a limited rollout budget.

\subsubsection{Impact of Error-turn Masking}

To verify the role of the error-turn masking mechanism in learning environmental feedback and error recovery capabilities, we further compare three SFT strategies:

\begin{itemize}
\item \textbf{Oracle Trajectory SFT:} Uses only manually verified, error-free expert trajectories for supervised learning. Here, the tool invocation code (action code) is manually checked to ensure the execution path is strictly correct, and the corresponding thought process is generated by the 32B teacher model. This setup is used to analyze the tool invocation capabilities the model can acquire when relying solely on idealized successful trajectory supervision.
\item \textbf{Unmasked SFT:} Uses teacher trajectories containing real "erroneous exploration $\rightarrow$ environmental feedback $\rightarrow$ strategy correction" processes, and computes the supervision loss on output tokens across all action spans. This forces the model to simultaneously learn correct execution behaviors and the erroneous actions generated during exploration.
\item \textbf{Masked SFT (Ours):} Retains complete error feedback information and subsequent recovery processes. However, for error action turns that cause environmental exceptions or violate task constraints, the supervision loss of the output tokens within the corresponding action span is zeroed out. This preserves supervision on all unmasked valid actions, including diagnostic, adjustment, and recovery behaviors following errors.
\end{itemize}

All three SFT strategies use the same 12 training tasks and the same number of training trajectories: 100 trajectories per task, totaling 1,200 trajectories for each strategy. Oracle Trajectory SFT uses error-free expert trajectories for these tasks, whereas Unmasked SFT and Masked SFT use the same teacher-generated successful trajectories containing failed attempts and recovery processes. The latter two settings differ only in whether the supervision loss on erroneous action turns is masked.

\begin{table}[!t]
\centering
\caption{Ablation Study of SFT Strategies for Agent Error Recovery Across Model Scales. Oracle, Unmasked, and Masked denote error-free expert trajectory SFT, full-trajectory SFT, and the proposed error-turn masking strategy, respectively.}
\label{tab:table2}
\scriptsize
\setlength{\tabcolsep}{2pt}
\renewcommand{\arraystretch}{1.05}
\resizebox{\linewidth}{!}{%
\begin{tabular}{llcccc}
\toprule
\textbf{Scale} & \textbf{SFT Strategy} & \textbf{Accuracy} & \textbf{Tool-Any-Order} & \textbf{Tool-In-Order} & \textbf{EIR} \\
\midrule
\multirow{4}{*}{4B} & Before Training & 31.8 & 47.6 & 40.1 & 45.8 \\
 & Oracle & 33.5 & 54.2 & 44.8 & 57.4 \\
 & Unmasked & 37.5 & \textbf{58.9} & 48.6 & 36.9 \\
 & Masked (Ours) & \textbf{41.5} & 58.6 & \textbf{49.4} & \textbf{31.6} \\
\midrule
\multirow{4}{*}{8B} & Before Training & 19.3 & 49.6 & 41.1 & 53.1 \\
 & Oracle & 35.2 & 56.7 & 47.5 & 57.8 \\
 & Unmasked & 44.9 & \textbf{62.5} & \textbf{52.9} & 23.7 \\
 & Masked (Ours) & \textbf{46.6} & 61.4 & 52.2 & \textbf{18.0} \\
\bottomrule
\end{tabular}}
\end{table}

As shown in Table~\ref{tab:table2}, all three SFT strategies can improve the execution capabilities of lightweight models in RS tool invocation tasks, indicating that supervised learning on agent interaction trajectories can effectively compensate for the shortcomings of base language models in tool use. For Qwen3-4B-Instruct, Accuracy increases from 31.8\% to 41.5\% after SFT; for Qwen3-8B, Accuracy increases from 19.3\% to 46.6\%. Simultaneously, both Tool-Any-Order and Tool-In-Order metrics improve, demonstrating that trajectory supervision not only enhances task completion capabilities but also improves the model's ability to model tool selection and execution workflows.

First, comparing Oracle Trajectory SFT with other failure-aware training strategies shows that merely learning ideal successful trajectories is insufficient to acquire robust environmental interaction capabilities. Although Oracle SFT improves tool matching metrics, its EIR conversely rises to 57.4\% and 57.8\%, indicating that solely relying on error-free trajectory supervision cannot improve the model's recovery capabilities when facing abnormal states. This suggests that error-free expert trajectories primarily provide execution patterns under ideal conditions but lack abnormal feedback and recovery processes from real environments. When the model enters unseen states or generates execution deviations, it remains unable to adjust its subsequent behaviors promptly due to a lack of error recovery experience. Therefore, for long-horizon tool invocation tasks, merely imitating correct paths fails to sufficiently learn state transition patterns in dynamic environments.

In contrast, after introducing successful trajectories containing failed attempts and recovery processes, Unmasked SFT can further reduce the proportion of error interactions and improve task completion capability. Taking Qwen3-4B-Instruct as an example, Unmasked SFT raises Accuracy from 33.5\% (Oracle SFT) to 37.5\%, while lowering EIR from 57.4\% to 36.9\%; on Qwen3-8B, Accuracy rises from 35.2\% to 44.9\%, and EIR drops from 57.8\% to 23.7\%. This result indicates that the environmental feedback and recovery processes contained in these successful trajectories can provide additional state transition information for the model. However, because Unmasked SFT supervises all action tokens, the model cannot distinguish between invalid erroneous actions generated during exploration and subsequent valid recovery behaviors, thus the erroneous actions themselves may also be learned by the model.

The Masked SFT proposed in this paper further resolves the aforementioned supervision noise problem. Experimental results show that Masked SFT achieves the best performance across both model scales. For Qwen3-4B-Instruct, Masked SFT improves Accuracy to 41.5\% and Tool-In-Order to 49.4\%, while reducing EIR to 31.6\%; for Qwen3-8B, Accuracy reaches 46.6\%, and EIR further drops to 18.0\%. Compared to Unmasked SFT, Masked SFT reduces EIR by 5.3 and 5.7 percentage points respectively, demonstrating that filtering out the supervision of erroneous actions can effectively reduce the model repeatedly generating invalid exploration behaviors.

It is worth noting that compared to Unmasked SFT, Masked SFT experiences slight declines in some tool matching metrics. For instance, on Qwen3-8B, Tool-Any-Order and Tool-In-Order drop by 1.1 and 0.7 percentage points, respectively. We believe this difference primarily stems from the distinct optimization objectives of the two supervision strategies: Unmasked SFT supervises complete interaction trajectories, which include erroneous actions generated during the exploration phase, allowing the model to learn tool invocation patterns closer to the sampled trajectories. Masked SFT, by masking out erroneous action spans, prevents the model from imitating invalid execution behaviors, shifting the learning focus toward state understanding and recovery processes after an error occurs. Therefore, Masked SFT may sacrifice a small amount of trajectory-level tool matching capability, but it can significantly reduce the EIR and enhance final task completion outcomes. This phenomenon illustrates that for long-horizon RS agents, simply optimizing tool invocation matching cannot fully measure agent capability; error recovery and environmental adaptability are equally critical factors determining task success rates.

Furthermore, Masked SFT does not simply delete failure experiences, but adopts an asymmetrical supervision strategy: erroneous actions are treated as low-quality supervision signals that need to be suppressed, while the environmental feedback and recovery behaviors following the error are retained as valid learning information. Consequently, the model can learn "how to recover from erroneous states" while avoiding learning "how to generate errors." This result verifies the core hypothesis of this paper: in complex RS agent scenarios, erroneous actions and environmental feedback in successful trajectories containing failed attempts and recovery processes hold different learning values. Erroneous actions may introduce supervision noise, while error feedback and subsequent recovery processes contain critical training signals required for the model to learn state transition patterns and recovery strategies. Furthermore, the consistent improvements observed on both Qwen3-4B-Instruct and Qwen3-8B demonstrate that the proposed error-turn masking strategy is not tied to a specific model scale, but provides a generally effective supervision scheme for learning robust tool-use behaviors under different model capacities.

\subsection{Ablation Study on Feedback-driven RL}

To verify the effectiveness of the two core designs in the RL phase---the MEFR and the TAP mechanism---we conducted component-level ablation experiments using the Qwen3-4B-Instruct model fine-tuned through our SFT phase as the base. We specifically compared the following five settings:

\begin{itemize}
\item \textbf{SFT Only:} Uses only the model after SFT, without RL optimization.
\item \textbf{Vanilla GRPO:} Adopts standard GRPO, providing binary rewards based purely on final task outcomes, without the MEFR and advantage protection.
\item \textbf{w/o MEFR:} Removes the MEFR, retaining only the binary outcome reward while keeping TAP, to analyze the role of reward shaping in exploration optimization.
\item \textbf{w/o TAP:} Retains MEFR but removes TAP, broadcasting the trajectory-level advantage indiscriminately to all action tokens.
\item \textbf{Ours:} Employs both the MEFR and TAP.
\end{itemize}

\begin{table}[!t]
\centering
\caption{Ablation Study of RL Components Based on the Qwen3-4B-Instruct Model After the Proposed SFT Phase.}
\label{tab:table3}
\scriptsize
\setlength{\tabcolsep}{2pt}
\renewcommand{\arraystretch}{1.05}
\resizebox{\linewidth}{!}{%
\begin{tabular}{lccccc}
\toprule
\textbf{Variant} & \textbf{Accuracy} & \textbf{Tool-Any-Order} & \textbf{Tool-In-Order} & \textbf{Avg Turns} & \textbf{EIR} \\
\midrule
SFT Only & 41.5 & 58.6 & 49.4 & 17.426 & 31.6 \\
Vanilla GRPO & 58.5 & 64.9 & 53.7 & \textbf{11.688} & 10.3 \\
w/o MEFR & 61.9 & 68.1 & 58.1 & 12.920 & 8.0 \\
w/o TAP & 63.1 & 68.6 & 58.6 & 13.068 & 7.7 \\
Ours & \textbf{65.9} & \textbf{69.1} & \textbf{58.9} & 12.761 & \textbf{6.1} \\
\bottomrule
\end{tabular}}
\end{table}

As shown in Table~\ref{tab:table3}, RL optimization can further enhance the RS task execution capabilities of models initialized via SFT. Compared to SFT Only, Vanilla GRPO improves Accuracy from 41.5\% to 58.5\%, Tool-Any-Order from 58.6\% to 64.9\%, while reducing EIR from 31.6\% to 10.3\%, and lowering Avg Turns from 17.426 to 11.688. This indicates that RL not only improves the task completion rate but can also reduce redundant exploration behaviors present during the SFT phase, driving the model to generate more compact tool invocation trajectories. Notably, Vanilla GRPO achieves the lowest Avg Turns (11.688), but its EIR remains at 10.3\%, which is significantly higher than our method's 6.1\%. This phenomenon suggests that when relying solely on final task outcome rewards, the model tends to learn shorter execution paths, but these paths do not necessarily correspond to more reliable environmental interaction strategies. Lacking process-level feedback, the model might avoid potential errors by reducing exploration turns, rather than learning how to effectively utilize environmental feedback to accomplish recovery. Therefore, simply reducing interaction length cannot fully reflect the execution efficiency of long-horizon agents; it must still be comprehensively evaluated alongside the EIR and task completion quality. Nevertheless, a clear gap remains between Vanilla GRPO and the subsequent improved methods, indicating that relying purely on sparse outcome rewards struggles to adequately guide the exploration process of complex, multi-turn RS agents.

When the MEFR is further removed from the complete method, model performance noticeably drops. \emph{w/o MEFR} reduces Accuracy from 65.9\% to 61.9\%, Tool-In-Order from 58.9\% to 58.1\%, while Avg Turns increases from 12.761 to 12.920. This result indicates that when depending strictly on final outcome rewards, although the model can learn some effective strategies, it struggles to sufficiently distinguish the contributions of behaviors at different stages during complex tool invocation processes. In RS tasks, some uncompleted trajectories may have already executed valid steps such as correct tool discovery, document understanding, or data flow progression, but receive only trajectory-level outcome supervision, which does not explicitly distinguish the contributions of these valid intermediate steps. By utilizing structured feedback inherently generated by the environment---including skill exploration progress, document acquisition status, and data flow advancement---this paper constructs process-level rewards, providing the model with more continuous policy optimization directions without the need to train extra evaluation models.

Further comparing \emph{w/o TAP} with the complete method allows us to analyze the role of the TAP mechanism. After removing this mechanism, the model still achieves relatively good performance (Accuracy 63.1\%, Tool-In-Order 58.6\%), showing that the MEFR already provides effective process supervision. However, compared to the complete method, the Accuracy in this setting drops by 2.8 percentage points, EIR increases from 6.1\% to 7.7\%, and Avg Turns increases from 12.761 to 13.068. This result indicates that when trajectory-level advantages are directly propagated to all action spans, some low-quality exploration behaviors may still receive update signals, thereby reducing the precision of policy optimization.

Ultimately, the complete method (Ours) achieves the best comprehensive performance. Compared to SFT Only, our method increases Accuracy from 41.5\% to 65.9\%, Tool-Any-Order from 58.6\% to 69.1\%, Tool-In-Order from 49.4\% to 58.9\%, while reducing EIR to 6.1\%. Compared to Vanilla GRPO, our method further boosts Accuracy by 7.4 percentage points and lowers the error interaction proportion by approximately 40.8\%. The MEFR and TAP play complementary roles: the former reduces invalid exploration by providing process-level signals related to task progression, enabling the model to complete tasks with fewer interaction turns; the latter further improves credit assignment within long-horizon trajectories, preventing error turns from receiving unreasonable updates, thereby simultaneously reducing the EIR and execution overhead.

Overall, this ablation experiment validates the effectiveness of the RL design in this paper. For RS agents requiring multi-turn tool invocations, solely relying on final task outcomes makes it difficult to fully utilize the information generated during interactions. By leveraging environmental feedback to construct process-level rewards and combining it with a turn-level advantage modulation mechanism, RL can far more fully utilize limited interaction data, thereby enhancing the autonomous execution capabilities of lightweight models in complex RS workflows.

\section{Discussion}

\subsection{Implications of Environment Feedback Learning}

The results of this paper reveal that environmental feedback can serve as more than transient execution information during agent inference; it can also function as a fundamental learning signal for capability evolution. Unlike static language modeling objectives that primarily optimize textual prediction, long-horizon RS agents must continuously adapt their behaviors according to dynamic environmental states, including tool execution results, exceptions, and intermediate data transitions.

Through RS-Claw-Evolution, environmental feedback is progressively utilized across different stages of agent capability development. During Experience Evolution, failure feedback enables lightweight agents to acquire error diagnosis and recovery behaviors beyond imitation of error-free expert trajectories. During Decision Evolution, environmental states and execution signals provide process-level optimization guidance, allowing RL to capture local decision quality beyond sparse terminal rewards.

These findings suggest that improving lightweight agent capability is not solely dependent on increasing model scale or expanding successful demonstrations. Instead, when training resources and task diversity are limited, environmental feedback provides additional state-transition information that helps models learn how actions influence future environments. Therefore, future RS agent development should move beyond learning static solutions and focus on learning adaptive interaction processes with continuous feedback.

\subsection{The Role of Imperfect Experiences in Agent Evolution}

For tool-driven agents, trajectory quality cannot be determined solely by final task success. Successful trajectories containing failed attempts and recovery processes can provide valuable information about environmental constraints, execution boundaries, and corrective strategies. Tool invocation errors, parameter mismatches, and data dependency failures represent important states that agents must handle in real-world deployment.

These trajectories contain components with different learning values. Erroneous actions themselves may introduce undesirable behavioral patterns, while the resulting environmental feedback and subsequent recovery processes provide critical information about state transitions and corrective strategies. Therefore, these trajectories should neither be discarded solely because they contain failed attempts nor be used for unselective behavioral imitation. Instead, their informative components should be selectively extracted for agent learning.

The proposed Error-turn Masking mechanism implements this principle by separating invalid actions from valuable recovery experiences. By masking supervision on erroneous action turns while preserving environmental feedback and subsequent correction behaviors, the model can learn to diagnose and recover from abnormal states without receiving direct supervision to imitate erroneous actions.

Experimental results further support this perspective. Compared with Oracle SFT trained only on error-free expert trajectories, failure-aware training reduces EIR, supporting the value of incorporating failure and recovery experiences into successful training trajectories. Compared with Unmasked SFT, which supervises all action turns in the same trajectories, Masked SFT further reduces EIR while improving task accuracy. These observations indicate that for long-horizon tool-use agents, the value of a trajectory should be determined not only by whether it succeeds, but also by the environmental information and behavioral adaptation strategies it contains.

\subsection{Limitations}

Although RS-Claw-Evolution effectively improves the execution capabilities of lightweight RS agents in long-horizon tool-use tasks, several limitations remain.

First, although this framework transforms environmental feedback into learning signals, the current implementation still partially relies on task-level reference workflows for constructing high-quality supervision. Specifically, reference code trajectories support hint generation during experience construction, while reference workflows and dataflows support process-level reward computation. While this design improves training efficiency and optimization stability in multi-stage RS analysis tasks, fully open-ended geospatial science exploration may not provide predefined execution paths or clearly specified intermediate targets and task constraints. Therefore, developing autonomous feedback-driven data generation and reward construction mechanisms by integrating environmental constraints, automated evaluation models, and semantic task understanding remains an important direction for future research.

Second, the current evaluation is conducted under a relatively stable RS tool ecosystem. In practical geospatial environments, tool libraries continuously evolve with the emergence of new RS platforms, geographic information systems (GIS), and specialized analysis libraries, introducing new interfaces and usage patterns. Enabling agents to continuously adapt to dynamic tool environments through tool understanding, online interaction, and continual learning, while preserving previously acquired tool-use capabilities, remains a critical challenge for long-term autonomous RS agents.

Finally, this work primarily evaluates RS-Claw-Evolution on the Earth-Bench benchmark. Although Earth-Bench covers diverse RS analysis workflows, its task distribution, tool ecosystem, and evaluation protocols cannot fully represent the complexity of real-world geospatial science scenarios. Future studies are needed to further validate the generalization ability of RS-Claw-Evolution across cross-platform RS systems, open-ended scientific exploration tasks, and broader tool-driven scientific computing domains.

\section{Conclusion and Future Work}

This paper proposes RS-Claw-Evolution, an environment-feedback-driven evolution framework for lightweight RS agents in long-horizon tasks. The framework focuses on training lightweight agents to exploit environmental interactions, execution feedback, and error signals for progressive capability improvement.
RS-Claw-Evolution progressively enhances lightweight agents through three evolutionary stages. First, the Interaction Evolution stage introduces a programming-based interaction paradigm that leverages executable code to explicitly manage intermediate states and environmental observations, reducing redundant context transmission during long-horizon workflows. Second, Experience Evolution uses hindsight analysis of failed interactions to guide the generation of successful trajectories containing failed attempts and recovery processes. Error-turn Masking excludes erroneous action turns from the direct supervised loss while retaining environmental feedback and recovery behaviors. Finally, the Decision Evolution stage introduces feedback-driven RL, where MEFR and TAP provide fine-grained optimization signals and improve credit assignment in long-horizon tool-use decision making.

Extensive experiments on the Earth-Bench RS agent benchmark demonstrate that RS-Claw-Evolution significantly improves lightweight models' capabilities in tool invocation, task planning, and error recovery using only a limited number of real RS tasks. Further analysis reveals that environmental feedback is not merely an execution-time observation signal, but can serve as a fundamental learning signal for agent capability evolution. By progressively learning interaction patterns, execution experiences, and decision strategies from environmental feedback, lightweight agents can achieve competitive performance with larger-scale models while maintaining efficient long-horizon interaction capabilities.

Future work will focus on developing more autonomous and adaptive environment-feedback-driven evolution mechanisms. First, we will explore self-supervised feedback learning strategies that reduce dependence on stronger teacher models for trajectory construction and on reference workflows for supervision and reward computation, enabling agents to operate in more open-ended geospatial science scenarios. Second, as RS software ecosystems continue to evolve, future research will investigate continual adaptation mechanisms that allow agents to acquire and integrate new tool capabilities through online interaction. Finally, we aim to extend RS-Claw-Evolution beyond RS toward broader scientific computing and tool-driven agent environments, investigating the applicability of feedback-driven agent evolution in diverse scientific domains.

\bibliographystyle{IEEEtran}
\bibliography{references}

\clearpage
\twocolumn[\begin{center}
{\LARGE Supplementary Material for RS-Claw-Evolution\par}
\vspace{1em}
\end{center}]
\setcounter{table}{0}
\setcounter{equation}{0}
\renewcommand{\theHtable}{supp.\arabic{table}}
\renewcommand{\theHequation}{supp.\arabic{equation}}
\lstset{breaklines=true,breakatwhitespace=false,basicstyle=\ttfamily\scriptsize,columns=fullflexible,keepspaces=true,frame=single,framesep=3pt,xleftmargin=2pt,xrightmargin=2pt}
\renewcommand{\thetable}{S\arabic{table}}
\renewcommand{\theequation}{S\arabic{equation}}

\appendices

\section{Dataset and Evaluation Details}

\subsection{Test Set Filtering}

Earth-Bench initially contains 248 remote sensing agent tasks. Since this work focuses on \textbf{interactive workflow execution with environment feedback}, we distinguish between workflow-based tool execution tasks and perception model invocation tasks.

We exclude the last 60 tasks (Task 189--248) from evaluation because they mainly involve perception-oriented tools, such as classification, object detection, visual grounding, counting, and segmentation. These tasks primarily retrieve pre-computed perception results from cached outputs rather than executing multi-step workflows with dynamic intermediate states and execution feedback.

In contrast, the first 188 tasks require agents to perform executable tool compositions, where actions produce intermediate artifacts and observations, including execution results, warnings, and error feedback. These characteristics align with our focus on feedback-driven learning for long-horizon tool-use agents.

Therefore, we retain the first 188 workflow execution tasks as the evaluation pool. Among them, 12 tasks are selected for training trajectory construction, while the remaining 176 tasks are reserved exclusively for testing.

\subsection{Training Task Selection}

To establish a low-resource learning setting, we select a small but diverse subset from the 188 executable workflows in Earth-Bench for training, while reserving the remaining tasks exclusively for evaluation. The goal is not to maximize the number of supervised tasks, but to expose the agent to diverse remote sensing concepts and workflow patterns, enabling the evaluation of generalization to unseen tasks.

The selected tasks are determined based on two criteria:

\textbf{(1) Domain diversity.} The training subset covers six major remote sensing application areas, including vegetation and water stress monitoring, land surface thermal environment, hydrology and cryosphere remote sensing, fire remote sensing, urban and human activity remote sensing, and atmospheric remote sensing.

\textbf{(2) Workflow diversity.} The selected tasks cover diverse reasoning structures, ranging from threshold-based analysis and temporal trend estimation to multi-step composite analysis. We avoid selecting tasks with highly overlapping execution patterns to reduce redundant supervision.

The final 12 training tasks and their workflow categories are listed in Table~\ref{tab:table4}.

\begin{table}[!t]
\centering
\caption{Selected Training Tasks.}
\label{tab:table4}
\scriptsize
\resizebox{\linewidth}{!}{%
\begin{tabular}{lll}
\toprule
\textbf{Task ID} & \textbf{Application Domain} & \textbf{Workflow Type} \\
\midrule
3 & Vegetation and Water Stress & Composite Analysis \\
43 & Vegetation and Water Stress & Threshold Analysis \\
14 & Land Surface Thermal & Temporal Trend \\
39 & Land Surface Thermal & Threshold Analysis \\
77 & Land Surface Thermal & Dual-object Difference \\
72 & Land Surface Thermal & Composite Analysis \\
93 & Land Surface Thermal & Dual-factor Analysis \\
18 & Atmospheric Remote Sensing & Temporal Trend \\
130 & Hydrology and Cryosphere & Dual-object Difference \\
170 & Hydrology and Cryosphere & Temporal Trend \\
181 & Fire Remote Sensing & Composite Analysis \\
187 & Urban and Human Activity & Temporal Trend \\
\bottomrule
\end{tabular}}
\end{table}

Basic statistical operations (e.g., mean, maximum, minimum, and variance computation) are not assigned dedicated training tasks because they frequently appear as intermediate operations in higher-level workflows and can be naturally learned through workflow-level supervision.

Overall, the training set contains only 12 out of 188 executable tasks (6.38\%), covering 28 out of 104 available tools (26.92\%). The remaining 176 tasks are reserved for evaluation, forming a low-resource generalization setting for assessing whether the learned agent can transfer workflow planning, tool selection, and environment interaction capabilities to unseen remote sensing tasks.

\subsection{Evaluation Metrics}

We follow the official Earth-Bench evaluation protocol and report both end-to-end task performance and tool execution metrics. In addition, to analyze the interaction reliability and context efficiency of remote sensing agents, we introduce additional trajectory-level and token-level metrics.

\subsubsection{End-to-End Metrics}

Following the Earth-Bench evaluator, we report two end-to-end metrics: Accuracy and Efficiency.

\textbf{Accuracy.} Accuracy measures the proportion of tasks where the final answer generated by the agent matches the Earth-Bench ground truth answer. The final answer is extracted from the agent response using the predefined answer tag patterns, including \texttt{\detokenize{<Answer>X<Answer>}} and \texttt{\detokenize{<Answer>X</Answer>}}. A task is considered successful if the extracted answer is consistent with the reference answer.

\textbf{Efficiency.} Efficiency measures the relative number of tool calls used by the agent compared with the reference execution trajectory:

\begin{equation}
\text{Efficiency}=\frac{N_{\text{model}}}{N_{\text{GT}}},
\label{eq:supp-1}
\end{equation}

where $N_{\text{model}}$ denotes the number of tool calls executed by the agent and $N_{\text{GT}}$ denotes the number of tool calls in the ground-truth trajectory. A value larger than 1 indicates that the agent performs more tool calls than the reference solution.

\subsubsection{Tool Execution Metrics}

Following the Earth-Bench evaluator, we measure the alignment between the agent's tool execution sequence and the ground-truth tool sequence using four step-level metrics.

\textbf{Tool-Any-Order.} This metric measures the proportion of ground-truth tools that appear in the predicted tool sequence regardless of their execution order:

\begin{equation}
\text{Tool-Any-Order}=\frac{\vert{}T_{\text{pred}}\cap T_{\text{GT}}\vert{}}{\vert{}T_{\text{GT}}\vert{}}.
\label{eq:supp-2}
\end{equation}

The tool sequences are treated as sets, where repeated tool invocations are removed.

\textbf{Tool-In-Order.} This metric evaluates whether the agent discovers the required tools following the correct execution order. Specifically, a greedy sequential matching strategy is adopted: for each ground-truth tool, the earliest unmatched occurrence in the remaining predicted sequence is selected. Intermediate irrelevant tool calls are allowed.

\textbf{Tool-Exact-Match.} This metric measures strict sequence-level alignment. The predicted and reference tool sequences are compared position by position, and matching continues until the first mismatch occurs or the shorter sequence terminates.

\textbf{Parameters.} This metric evaluates whether the predicted tool calls contain exactly the same input parameters as the reference execution. The comparison is performed step by step and terminates when the first unmatched tool call or parameter mismatch occurs.

All tool execution metrics use soft scoring, where each task receives a partial score based on the matched proportion, and the final metric is averaged over all evaluation tasks.

\subsubsection{Interaction Reliability Metrics}

To analyze the agent's ability to handle environment feedback and recover from execution failures, we introduce two trajectory-level metrics.

\textbf{Error Interaction Rate (EIR).} EIR measures the proportion of interaction turns that involve explicit execution errors or invalid interactions during the entire trajectory. Unlike counting individual error labels, EIR evaluates the occurrence of erroneous interaction states at the turn level, avoiding over-counting when a single environment response contains multiple error indicators (e.g., execution failure, runtime exception, and status failure).

For a trajectory $\tau_i$ containing $T_i$ assistant action turns, we first identify the set of error turns:

\begin{equation}
\mathcal{E}_i = \{t \mid \mathcal{E}(a_{i,t}, o_{i,t}) = 1\}
\label{eq:supp-3}
\end{equation}

where $\mathcal{E}(a_{i,t}, o_{i,t})$ is the error indicator defined in the Task Formalization subsection of the main paper, which equals 1 when the action produces an explicit environment error, violates execution constraints, or triggers an invalid interaction state. The trajectory-level error interaction ratio is defined as:

\begin{equation}
\text{EIR}_i = \frac{\vert{}\mathcal{E}_i\vert{}}{T_i}
\label{eq:supp-4}
\end{equation}

The final EIR is averaged over all evaluation tasks:

\begin{equation}
\text{EIR} = \frac{1}{N}\sum_{i=1}^{N}\text{EIR}_i
\label{eq:supp-5}
\end{equation}

A lower EIR indicates that the agent requires fewer erroneous interactions during task execution, reflecting stronger interaction reliability and error avoidance capability.

\textbf{Average Turns (Avg Turns).} Avg Turns measures the average number of assistant action turns required to complete each task. Only generated action blocks are counted, excluding environment responses:

\begin{equation}
\text{Avg Turns}=\frac{1}{N}\sum_{i=1}^N T_i.
\label{eq:supp-6}
\end{equation}

\subsubsection{Context Efficiency Metrics}

To quantify the reduction of interaction overhead introduced by the programming-based action representation, we additionally measure the token consumption of actions and observations.

\textbf{Action Tokens.} Action Tokens denote the average number of tokens generated by the agent's action blocks per task. Only executable action contents (e.g., generated Python code) are counted.

\textbf{Observation Tokens.} Observation Tokens denote the average number of tokens returned by the environment per task. This includes tool execution outputs, error tracebacks, and warning messages:

\begin{equation}
\text{Observation Tokens}=\frac{1}{N}\sum_{i=1}^N O_i.
\label{eq:supp-7}
\end{equation}

These token-level metrics are used to analyze how different interaction representations affect the context burden during long-horizon remote sensing workflows.

\section{Experimental Settings and Computational Cost}

\subsection{SFT Settings}

During the SFT phase, we train the Qwen3-4B-Instruct and Qwen3-8B models. The training dataset for each SFT strategy contains 1,200 trajectories from the same 12 selected tasks, with 100 trajectories per task. Oracle Trajectory SFT uses error-free expert trajectories, while Unmasked SFT and Masked SFT use the same teacher-generated successful trajectories containing failed attempts and recovery processes. We train both models for 2 epochs using a global batch size of 4. The learning rate is initialized to $5 \times 10^{-6}$ and follows a cosine decay schedule down to a minimum of $5 \times 10^{-7}$, with a warmup fraction of 0.1. Detailed hyperparameters for the SFT phase are listed in Table~\ref{tab:table5}.

\begin{table}[!t]
\centering
\caption{Hyperparameters for Supervised Fine-Tuning (SFT).}
\label{tab:table5}
\scriptsize
\resizebox{\linewidth}{!}{%
\begin{tabular}{ll}
\toprule
\textbf{Hyperparameter} & \textbf{Value} \\
\midrule
Training epochs & 2 \\
Global batch size & 4 \\
Max tokens per GPU & 20,480 \\
Max sample tokens & 20,480 \\
Learning rate & $5 \times 10^{-6}$ \\
Minimum learning rate & $5 \times 10^{-7}$ \\
LR schedule & Cosine decay \\
Warmup fraction & 0.1 \\
Optimizer & Adam ($\beta_1=0.9, \beta_2=0.95$) \\
Weight decay & 0.1 \\
\bottomrule
\end{tabular}}
\end{table}

\subsection{RL Settings}

In the RL phase, we employ Group Relative Policy Optimization (GRPO) to optimize the exploration policy. For each prompt, we sample a group of 12 rollouts (group size $G = 12$) to estimate the advantage. The reward function parameters are empirically set to balance task completion and efficiency:

\begin{itemize}
\item \textbf{Core Task Incentives (}$\alpha_{\text{success}} = 10, \alpha_{\text{fail}} = -1.0$\textbf{):} The base success reward is significantly higher than the absolute value of the failure penalty to encourage the model to engage in active policy exploration. The sparse success reward at the end of the episode far exceeds the minor trajectory penalties, thereby providing a clear directional gradient for policy optimization.
\item \textbf{Trade-off Between Efficiency and Correctness (}$\lambda = 0.1, \mu = 0.25$\textbf{):} The turn penalty ($\lambda = 0.1$) serves as a mild regularization term used to suppress infinite loops or redundant tool invocations without overly penalizing necessary long-horizon reasoning. In contrast, the error penalty ($\mu = 0.25$) is deliberately set higher than the single-turn cost to severely penalize syntax or execution errors, forcing the model to favor safe and robust actions.
\item \textbf{Dense Process Guidance (}$\omega_{\text{data}} = 1.0, \omega_{\text{doc}} = 0.25, \omega_{\text{skill}} = 0.2$\textbf{):} To alleviate the reward sparsity problem in long-horizon tasks, process rewards provide intermediate reward shaping. The essence of remote sensing workflows is "dataflow"---passing the correct files to the correct algorithms constitutes substantive progress; therefore, $\omega_{\text{data}} = 1.0$. In contrast, viewing the skill tree and consulting API documents are merely meta-operations or auxiliary explorations. They are assigned supplementary weights to encourage appropriate tool exploration without overshadowing the core execution objective.
\end{itemize}

The RL training runs for 5 epochs with a constant learning rate of $1 \times 10^{-6}$. Detailed configuration for GRPO and rollouts is summarized in Table~\ref{tab:table6}.

\begin{table}[!t]
\centering
\caption{Hyperparameters for Reinforcement Learning (GRPO).}
\label{tab:table6}
\scriptsize
\resizebox{\linewidth}{!}{%
\begin{tabular}{ll}
\toprule
\textbf{Hyperparameter} & \textbf{Value} \\
\midrule
Training epochs & 5 \\
Learning rate & $1 \times 10^{-6}$ \\
LR schedule & Constant \\
Optimizer & Adam ($\beta_1=0.9, \beta_2=0.98$) \\
Weight decay & 0.1 \\
KL loss coefficient & 0.001 \\
KL loss type & k2 \\
Entropy coefficient & 0 \\
PPO/GRPO clipping & $\epsilon=0.2, \epsilon_{\text{high}}=0.28$ \\
Rollout group size & 12 \\
Rollout temperature & 1 \\
Max response length & 2,048 \\
Training context length & 14,336 \\
\bottomrule
\end{tabular}}
\end{table}

\subsection{Computational Cost Analysis}

We quantify the computational overhead and time costs across the three main stages of our framework:

\begin{enumerate}
\item \textbf{Trajectory Generation Cost:} The hint-guided teacher rejection sampling is performed using the Qwen3-32B model with a temperature of 1.0 and a top-$p$ value of 0.95. The trajectory collection process is executed on 2 NVIDIA A100 GPUs and takes approximately 39 hours.
\item \textbf{SFT Cost:} The SFT process is accelerated using 2 NVIDIA A100 GPUs. Training the Qwen3-4B-Instruct model takes around 3 hours, while training the Qwen3-8B model requires 5 hours and 47 minutes.
\item \textbf{RL Cost:} The GRPO training phase utilizes 4 NVIDIA A100 GPUs. Completing 5 epochs of reinforcement learning for the Qwen3-4B-Instruct model takes approximately 6 hours.
\end{enumerate}

To further illustrate the Masked SFT strategy proposed in this paper, we detail the criteria used to identify invalid tool invocation turns and present the specific statistical distribution of this masking mechanism on the training dataset.

\section{Error-turn Masking Analysis}

\subsection{Mask Criterion}

When the agent generates explicit exceptions during Python execution or tool invocation, or violates the tool invocation and observation disclosure policies, the turn is judged as an "error turn" and undergoes loss masking. These error states are automatically logged by the interactive environment during execution through a two-stage verification process:

\begin{itemize}
\item \textbf{Pre-execution Checking:} Before executing the code block output by the model, the environment first conducts policy verification on it. For example, it checks whether the progressive disclosure order has been violated, whether exactly the same code has been repeatedly submitted, or whether invalid actions have been repeatedly executed. If it fails the check, the environment will intercept the operation, generate a \texttt{\detokenize{system_inject}} or \texttt{\detokenize{invalid_action_warning}} record, and write the specific warning pattern into the log, instead of executing it as a normal tool invocation.
\item \textbf{Runtime Execution Capture:} If the code passes the pre-execution check, the environment will run the code in a controlled Python namespace. The executor precisely captures general Python exceptions, exceptions thrown inside tool functions, and behaviors of repeatedly passing the same runtime parameters to \texttt{\detokenize{earth_agent.call(...)}}, recording them as \texttt{\detokenize{exec_result["error"]}} and writing them into the \texttt{\detokenize{code_execution.error}} field. Simultaneously, the tool wrapper writes a runtime audit for each invocation: a successful invocation is recorded as \texttt{\detokenize{status="ok"}}, an internal tool exception is recorded as \texttt{\detokenize{status="error"}}, and an intercepted repetition of the same parameters is recorded as \texttt{\detokenize{status="blocked_repeated_call"}}.
\end{itemize}

Therefore, the structured state fields directly from the real executor and tool wrapper constitute the precise basis for the masking operation.

\subsection{Mask Ratio Statistics}

The SFT training set contains 1,200 trajectories from 12 tasks, with 100 trajectories per task. The masking statistics below are computed over this complete training set.

To quantify the impact of the error-turn masking mechanism, we calculated the mask ratios at both the turn level and the token level. The token-level statistics utilized the exact same Qwen3 tokenizer, chat template, and \texttt{\detokenize{loss-mask-type=qwen3}} settings as the actual training. The specific statistical data is shown in Table~\ref{tab:table7}.

\begin{table}[!t]
\centering
\caption{Turn-Level and Token-Level Masking Statistics.}
\label{tab:table7}
\scriptsize
\resizebox{\linewidth}{!}{%
\begin{tabular}{lll}
\toprule
\textbf{Metric} & \textbf{Unmasked SFT} & \textbf{Masked SFT (Ours)} \\
\midrule
\multicolumn{3}{l}{\textbf{Turn-level Statistics}} \\
Total Agent Turns & 17,816 & 17,816 \\
Trainable Turns & 17,816 & 14,929 \\
Masked Turns & 0 & 2,887 \\
Turn Retention Ratio & \textbf{100\%} & \textbf{83.80\%} \\
Turn Masking Ratio & \textbf{0\%} & \textbf{16.20\%} \\
\midrule
\multicolumn{3}{l}{\textbf{Token-level Statistics}} \\
Total Trainable Tokens & 3,143,149 & 2,603,950 \\
Masked-out Tokens & 0 & 539,199 \\
Token Retention Ratio & \textbf{100\%} & \textbf{82.85\%} \\
Token Masking Ratio & \textbf{0\%} & \textbf{17.15\%} \\
\bottomrule
\end{tabular}}
\end{table}

The results indicate that the masking mechanism effectively filters out approximately 17.15\% of the supervision tokens corresponding to invalid exploration and noise in the training set, enabling the model to focus on learning error recovery and valid reasoning paths.

\section{Reward Computation}

To provide dense supervision signals during the Reinforcement Learning (RL) phase, we dynamically compute multi-dimensional environment feedback rewards based on the agent's intermediate environmental interactions. This section details the computation methods for the Data-flow Progress Reward and the Documentation Progress Reward.

\subsection{Data-flow Progress Reward}

The reference data flow is generated from the ground-truth trajectories provided by Earth-Bench, constructed by extracting input and output data families from function parameters, keyword arguments, assignment targets, and path strings.

The matching of data families is not a mechanical string comparison of complete paths, but rather a normalization of paths, filenames, wildcards, variable names, and the provenance information of tool outputs. For example, parameters such as \texttt{\detokenize{output_path}}, \texttt{\detokenize{output_paths}}, \texttt{\detokenize{save_name}}, and \texttt{\detokenize{save_path}} are uniformly treated as output locations. Crucially, if the output of a certain tool is used as the input for a subsequent tool, the system records a specific provenance tag (e.g., a provenance token indicating "this input comes from the output of an upstream tool"). This mechanism strictly distinguishes between two fundamentally different situations: "using the correct tool" and "passing the wrong file to the correct tool."

The data flow of the actual trajectory is dynamically constructed from runtime tool invocation audit records. The system only processes successfully executed \texttt{\detokenize{earth_agent.call}} records, and extracts runtime input/output data families and provenance information from each successful invocation. Failed tool invocations do not yield any data-flow progress reward because they fail to generate reliable intermediate results.

During matching, the system groups the reference steps by tool name. A reference tool node is judged as "complete" only when all of the following conditions are simultaneously met:

\begin{enumerate}
\item The corresponding tool is successfully invoked in the trajectory;
\item The runtime inputs cover the input data families required by the reference node or match the upstream provenance information;
\item The runtime outputs cover the output data families required by the reference node, and the number of generated outputs is no less than the requirement of that tool group.
\end{enumerate}

The calculation process of this reward can be expressed as the following pseudocode:

\begin{codepanel}[frametitle={Algorithm 1: Data-flow Progress Reward}]
\begin{lstlisting}[frame=none,aboveskip=0pt,belowskip=0pt,xleftmargin=0pt,xrightmargin=0pt]
Input: reference trajectory T*, runtime audit log L
Output: data-flow progress reward r_flow
1:  if T* contains annotated data-flow steps then
2:      G <- load annotated reference dependency nodes from T*
3:  else
4:      G <- extract reference dependency nodes from the AST of ground-truth code
5:  end if

6:  C <- empty ordered runtime context

7:  for each record l in L in chronological order do
8:      if l.status != "ok" or l.tool_name is empty then
9:          continue
10:     end if
11:     x <- normalize input families, output families, paths, and provenance in l
12:     append x into C
13: end for

14: completed <- 0

15: for each reference dependency node g in G do
16:     matched, reason <- CompleteDataflowNode(g, C)
17:     append expected fields, runtime fields, and reason into match_trace
18:     if matched then
19:         completed <- completed + 1
20:     end if
21: end for

22: r_flow <- completed / max(|G|, 1)

23: return r_flow
\end{lstlisting}
\end{codepanel}

where \texttt{\detokenize{CompleteDataflowNode}} is used to determine whether the runtime tool invocation completes the corresponding node in the reference dataflow:

\begin{codepanel}[frametitle={Algorithm 1a: CompleteDataflowNode}]
\begin{lstlisting}[frame=none,aboveskip=0pt,belowskip=0pt,xleftmargin=0pt,xrightmargin=0pt]
Input: reference node g, runtime context C
Output: matched flag

1: Extract required input families, sources, and provenance constraints from g
2: Extract runtime input/output families and provenance from C

3: if provenance constraints exist then
4:     verify provenance coverage
5: else if source constraints exist then
6:     verify source and family coverage
7: else
8:     verify input family coverage
9: end if

10: verify required output families and output cardinality

11: return True if all constraints are satisfied
\end{lstlisting}
\end{codepanel}

\subsection{Tool Discovery Progress Reward}

For each question, the system first extracts the target skill set $\mathcal{S}$ and target document set $\mathcal{D}$ based on the reference tool sequence and the tool registry. Here, a \emph{skill} represents the toolkit to which the tool belongs (e.g., \texttt{\detokenize{inversion}}), while a \emph{doc} represents specific tool documentation (e.g., \texttt{\detokenize{inversion.modis_day_night_lst}}).

During the agent's interactive runtime, the system parses the Abstract Syntax Tree (AST) of every generated Python code block and specifically captures explicit EarthAgent disclosure actions:

\begin{lstlisting}
earth_agent.skill("kit")
earth_agent.doc("kit", "tool_name")
earth_agent.call("kit", "tool_name", ...)
\end{lstlisting}

The toolkit name and tool name here must be parsable from the AST as literal strings. If the model claims "I should check the inversion tool" in its natural language planning, it will not be counted towards document progress; tool names written in code comments are equally invalid.

The calculation process of this reward can be expressed as the following pseudocode:

\begin{codepanel}[frametitle={Algorithm 2: Documentation Progress Reward}]
\begin{lstlisting}[frame=none,aboveskip=0pt,belowskip=0pt,xleftmargin=0pt,xrightmargin=0pt]
Input: reference tool sequence P*, tool registry M, executed code blocks B
Output: skill progress reward r_skill, documentation progress reward r_doc
1:  S* <- empty set
2:  D* <- empty set

3:  for each tool p in P* do
4:      kit <- lookup toolkit of p in M
5:      S* <- S* union {kit}
6:      D* <- D* union {(kit, p.name)}
7:  end for

8:  seen_skills <- empty set
9:  seen_docs <- empty set

10: for each code block b in B in chronological order do
11:     A <- parse b with AST and extract skill/doc actions in source order

12:     execute b in the environment
13:     if b raises an exception then
14:         continue
15:     end if

16:     for each action a in A do
17:         valid, kind, key <- ValidDocumentationAction(a, seen_skills)

18:         if not valid then
19:             continue
20:         end if

21:         if kind == "skill" then
22:             seen_skills <- seen_skills union {key}
23:         end if

24:         if kind == "doc" then
25:             seen_docs <- seen_docs union {key}
26:         end if
27:     end for
28: end for

29: r_skill <- |seen_skills intersection S*| / max(|S*|, 1)
30: r_doc <- |seen_docs intersection D*| / max(|D*|, 1)

31: return r_skill, r_doc
\end{lstlisting}
\end{codepanel}

where \texttt{\detokenize{ValidDocumentationAction}} evaluates whether the exploration action parsed from the AST is valid: only when the code block containing these exploration actions executes successfully without throwing exceptions are the corresponding actions added to the \texttt{\detokenize{seen_skill_kits}} and \texttt{\detokenize{seen_tool_docs}} sets, respectively. If a code block fails to execute, the document discovery actions within it are not considered reliable progress.

\section{System Prompts \& Interaction Formats}

\subsection{RS-Claw CodeAct Interaction Prompt}

To achieve robust environmental interaction, we adopt a CodeAct-style prompt, instructing the agent to directly write executable Python code rather than static JSON formats. The base prompt structure is as follows:

\begin{codepanel}[frametitle={CodeAct Interaction Prompt}]
\begin{lstlisting}[frame=none,aboveskip=0pt,belowskip=0pt,xleftmargin=0pt,xrightmargin=0pt]
# Role & Task
You are an Earth Scientist answering Earth Observation (EO) related multiple-choice questions by writing Python code to call remote sensing tools (EO tools). Please plan your steps and pacing reasonably.
---
# EO Tool System
## earth_agent API
{api_table}
## Kit Description
{kit_table}
---
Interaction Specifications (CodeAct)
Output strictly in the following format for each turn:
**Thought**: Analyze the current state. In the first turn, use **[Plan]** to analyze the question and plan the execution strategy.
In subsequent turns, you MUST use concise, telegraphic style and strictly follow this micro-structure. Narrating history or generating verbose paragraphs is strictly prohibited:
**[Diagnosis]**: Stating the core result of the Observation or the root cause of an error. (Ensure brevity)
**[Next]**: Stating the exact tool or code logic to be used next. (Be concise)
**Action**: Your action in each turn **MUST ONLY** output **ONE** of the following: [Execution Code] OR [Final Answer]:
1. **Execution Code**: Provide a Python code block (```python ... ```) to execute the operation decided in Thought. The system will execute it and return the stdout/stderr as **Observation**, entering the next turn.
2. **Final Answer**: When the calculation is complete and no more code execution is needed, provide the answer in an independent reply without any code blocks to end the task.
## Tool Invocation Constraints
- **Strictly follow skill -> doc -> call order**: You cannot call `earth_agent.doc` before calling `earth_agent.skill`; and you cannot call `earth_agent.call` before calling `earth_agent.doc`. Additionally, `earth_agent.skill`, `earth_agent.doc`, and `earth_agent.call` must be executed in separate code blocks. Parameter names must be copied verbatim from the `doc` documentation, and tool names must be copied verbatim from the `skill` list. No guessing or fabrication is allowed.
- **Process all data**: When involving multiple files or time series, must use for loop to process all files, not just one sample; if tool only accepts single file path (not list), should call tool separately for each file and collect results, not process only partial files citing "tool doesn't support batch"
- **Intermediate output paths**: Use short relative filenames only for output parameters such as `output_path`. For downstream inputs, store and pass the actual path returned by `earth_agent.call()`; never reuse the relative filename you supplied as an input path.
## Core Runtime Specifications
- **Global Object Injections**: `earth_agent` is a globally injected built-in object provided by the runtime ecosystem, NOT a standard Python module.
  - **CRITICAL**: Absolutely NO `import earth_agent` or `from earth_agent import ...` statements allowed. 
  - **Usage**: Access all tools and methods directly through the global `earth_agent` variable (e.g., `earth_agent.skill("statistics")`).
## Coding Specifications
- **Strict Two-Step Initialization**: In the very first execution turn, you are ONLY allowed to retrieve the file list and count using `earth_agent.filelist()`. ANY other operations (such as parsing filenames, filtering, or initializing downstream logic) are STRICTLY FORBIDDEN until the file names are actually observed in the next turn.
- Each response contains only one code block, code block must be preceded by Thought; system only executes first code block, subsequent code blocks and self-simulated Observations have no effect
- Code must use `print()` to output results, otherwise cannot see Observation
- All code blocks within same question share namespace, variables can be used across blocks
- Can infer time, data type, etc. from filename strings, flexibly use Python string operations to extract needed content
- Can use basic Python statements (variable assignment, list comprehension, string slicing, sorting, etc.) to assist data processing, no need to call earth_agent
- **Filter files using variable operations**: Filter the list returned by `earth_agent.filelist()`, strictly forbidden to hardcode filenames into code
- **Immediately switch approach on error**: If code errors, next round's Thought must analyze error root cause (tool doesn't exist? Parameter name wrong? Path format wrong?), then adopt completely different method; forbidden to repeatedly submit same code without any modifications
- **Diagnose before acting on abnormal results**: If tool returns empty list, all-zero sequence, or results obviously inconsistent with expectations, must diagnose step-by-step in Thought (are filter conditions too strict? Do files exist? Is data valid?), then adjust code after clear diagnosis; must not repeatedly call same code that produces abnormal results
## Answer Format
- **Before giving answer**: Must explicitly write computed result values in Thought
- **Answer format**: Provide `<Answer>option</Answer>` in standalone response without code block; strictly forbidden to give answer in same response containing code block
\end{lstlisting}
\end{codepanel}

The specific API definition injected into the \texttt{\detokenize{{api_table}}} placeholder is as follows:

\begin{codepanel}[frametitle={CodeAct API Definitions}]
\begin{lstlisting}[frame=none,aboveskip=0pt,belowskip=0pt,xleftmargin=0pt,xrightmargin=0pt]
[
  {
    "method": "earth_agent.skill(kit)",
    "desc": "Read Skill documentation for the specified kit to get names and descriptions of all available tools in that kit. Call this tool before calling tool_doc or call_eo_tool to learn which tools are available in the kit.",
    "params": [
      {
        "name": "kit",
        "type": "str",
        "desc": "Kit name, options: index/inversion/perception/analysis/statistics"
      }
    ],
    "returns": "str: Complete tool list documentation for the kit in Markdown format"
  },
  {
    "method": "earth_agent.doc(kit, tool_name)",
    "desc": "View detailed documentation for an EO tool, including parameter descriptions and return values.",
    "params": [
      {
        "name": "kit",
        "type": "str",
        "desc": "Kit name, options: index/inversion/perception/analysis/statistics"
      },
      {
        "name": "tool_name",
        "type": "str",
        "desc": "Tool function name"
      }
    ],
    "returns": "str: Complete documentation for the tool"
  },
  {
    "method": "earth_agent.call(kit, tool_name, **kwargs)",
    "desc": "Call an EO tool to perform Earth observation computation.",
    "params": [
      {
        "name": "kit",
        "type": "str",
        "desc": "Kit name, options: index/inversion/perception/analysis/statistics"
      },
      {
        "name": "tool_name",
        "type": "str",
        "desc": "Tool function name"
      },
      {
        "name": "**kwargs",
        "type": "dict",
        "desc": "Tool-specific keyword arguments copied from earth_agent.doc output"
      }
    ],
    "returns": "Tool execution result, such as numeric values, dictionaries, lists, or saved file paths"
  },
  {
    "method": "earth_agent.filelist(dir_path)",
    "desc": "Return a list of complete file paths in the specified directory.",
    "params": [
      {
        "name": "dir_path",
        "type": "str",
        "desc": "Directory path, e.g., \"benchmark/data/question2\""
      }
    ],
    "returns": "list[str]: Complete path for each file in the directory, excluding hidden files"
  }
]
\end{lstlisting}
\end{codepanel}

Additionally, the skill tree structure remains consistent with RS-Claw.

\subsection{RS-Claw JSON Interaction Prompt}

In the comparative experiments, we adopt the standard JSON-formatted interaction prompt from the original RS-Claw paper.

\begin{codepanel}[frametitle={JSON Interaction Prompt}]
\begin{lstlisting}[frame=none,aboveskip=0pt,belowskip=0pt,xleftmargin=0pt,xrightmargin=0pt]
You are a geoscientist, and you need to use tools to answer multiple-choice
questions about Earth observation data analysis. Note that if a tool returns
an error, you can only try again once. Ultimately, you only need to explicitly
tell me the correct choice.

ATTENTION:
1. When a tool returns "Result saved at /path/to/file", you must use the full
   returned path "/path/to/file" in all subsequent tool calls.
2. For each question, you must provide the choice you think is most appropriate.
   Your final answer format must be:
   <Answer>Your choice</Answer>
3. You have access to EO tools via a 4-meta-tool progressive disclosure
   interface: skill, doc, call, filelist.
   You MUST follow the strict order: skill -> doc -> call.
   Skipping any step is forbidden.
   If a tool call returns an error, re-read the doc output carefully and fix
   the parameters. Only retry once --- if it fails again, try a different tool
   or approach.
   For doc and call, use tool_id in 'kit.tool_name' format,
   e.g. 'statistics.calc_batch_image_mean'.
   For call, tool_args must be a JSON string,
   e.g. tool_args='{"file_list": ["a.tif", "b.tif"]}'
   Available kits and their applicable tasks:
{kit_table}
\end{lstlisting}
\end{codepanel}

\subsection{Hint Generation Prompt}

This prompt is utilized during SFT data construction. It takes the task description, ground-truth code trajectory, relevant tool descriptions, and failed teacher-agent trajectories as input. The hint generator analyzes discrepancies between failed interactions and successful reference workflows to extract corrective guidance. The resulting hint does not directly provide a complete executable solution or a specific sequence of domain-tool calls, but retains necessary control logic and task constraints, including iteration strategies, data dependencies, and parameter requirements. Concrete tool discovery and invocation remain the responsibility of the teacher agent.

\begin{codepanel}[frametitle={Hindsight Hint Generation Prompt}]
\begin{lstlisting}[frame=none,aboveskip=0pt,belowskip=0pt,xleftmargin=0pt,xrightmargin=0pt]
You are an expert analyst for remote sensing autonomous agents, specializing in multi-tool geospatial reasoning and execution planning.

Your objective is to perform hindsight analysis on a failed teacher-agent trajectory and generate a compact, actionable execution hint that helps an agent successfully solve the same remote sensing task.

The generated hint should capture transferable execution knowledge rather than reproduce the original solution. It should explain how to avoid previous failures, discover required capabilities, correctly process remote sensing data, and complete the task through progressive tool disclosure.

You are provided with three types of information:

(1) Failed Execution Trajectory:
A teacher agent's unsuccessful interaction history, including observations, actions, tool exploration attempts, execution outcomes, errors, warnings, and intermediate reasoning.

(2) Ground-Truth Tools and Toolkits:
The names, functional descriptions, input requirements of the tools used in the successful execution, together with their hierarchical toolkit (kit) organization.

(3) Ground-Truth Code (gt_code):
The complete successful Python execution code. It provides a reference execution workflow, including high-level data dependencies, interaction structure, and task-specific execution constraints.

Your task:

First, analyze the failed trajectory to identify the key causes of failure, such as:
- incorrect assumptions about available remote sensing data,
- improper tool discovery or toolkit navigation,
- missing prerequisite documentation reading,
- incorrect temporal/spatial data matching,
- invalid filtering strategies,
- incorrect parameter configuration,
- incomplete recovery strategies.

Then, compare the failure patterns with the successful reference execution to derive corrective guidance, required capabilities, and interaction strategies.

The final hint should summarize the corrective knowledge needed for a agent to successfully execute the task.

---

### Critical Constraint: Progressive Disclosure

Tool usage follows a strict hierarchical progressive disclosure mechanism:

1. The agent MUST first discover and read the documentation of the required toolkit (kit).
2. Only after accessing the toolkit can the agent discover the available tools inside it.
3. The agent MUST read the documentation of a specific tool before invoking that tool.

Therefore:

- The generated hint MUST explicitly instruct the agent to search/read the required toolkit by its exact toolkit name before describing the capability needed from that toolkit.
- Toolkit names may be explicitly mentioned.
- Specific tool names, API names, or function names from the ground-truth solution MUST NOT be mentioned.
- When referring to tools, describe only their functional capability, expected inputs, and outputs.

Example:
Correct:
"Search the `inversion` kit for a capability that performs day/night thermal product inversion using paired raster inputs."

Incorrect:
"Use `xxx_lst_tool`."

---

### Remote Sensing Task Awareness

The generated hint should preserve remote sensing-specific execution knowledge whenever relevant, including:

- Temporal consistency:
  Preserve important temporal iteration rules, acquisition ordering, date formatting, and cross-product temporal alignment requirements.

- Data discovery:
  Encourage inspecting available remote sensing files rather than assuming fixed filenames, timestamps, orbit information, or acquisition conditions.

- Multi-product alignment:
  Preserve necessary relationships among different remote sensing products, including spatial compatibility, temporal matching, and semantic correspondence.

- Raster processing logic:
  Retain essential operations involving raster generation, filtering, threshold-based analysis, and spatial statistics.

- Remote sensing pitfalls:
  Highlight failure-prone assumptions caused by remote sensing data characteristics, such as variable acquisition times, missing observations, product naming inconsistencies, or incomplete daily coverage.

Do not introduce general remote sensing background knowledge. Only include domain-specific constraints required for successfully completing the task.

---

### The generated hint MUST NOT:

- Directly provide the final answer.
- Mention any specific tool names, API names, or function names from the ground-truth solution.
- Reveal the exact implementation code.
- Blindly translate the entire gt_code into natural language.
- Describe generic background knowledge unrelated to task execution.
- Skip the required toolkit discovery step.
- Simply say "call a tool" without explaining what capability should be searched and why.
- Expose or reference any relative or absolute file paths (e.g.,"question14/lst_{timestamp}.tif") in the generated hint.

---

### The generated hint SHOULD:

- Prioritize mistakes and missing knowledge observed from the failed trajectory.
- Explain how to avoid brittle strategies that caused failure.
- Extract transferable execution patterns from the successful solution.
- Preserve important control logic from the successful execution, such as:
  - iteration strategies,
  - temporal matching rules,
  - string filtering logic,
  - data validation conditions,
  - indexing or ordering requirements,
  - parameter constraints,
  - final result interpretation logic.

- Describe required tools through their capabilities rather than names.
- Mention the corresponding toolkit before describing the required capability.
- Follow the actual execution order of the task.
- Be concise and operational, suitable as an additional hint provided to a agent during inference.

---

### Output Format Requirement:

Generate ONLY the execution hint.

The hint MUST:
- Be completely enclosed within `<hint>` and `</hint>` tags.
- Contain only a numbered list of execution steps.
- Each step should describe one actionable instruction.
- Avoid any JSON formatting, explanations, introductions, or markdown code blocks.

Example output format:

<hint>
1. ...
2. ...
3. ...
</hint>

---

Input:
Task:
{task}

Failed trajectory:
{failed_trajectory}

Ground-truth tools and toolkits:
{ground_truth_tools_and_kits}

Ground-truth code:
{gt_code}

Output:
Generate the hint enclosed in <hint></hint>.
\end{lstlisting}
\end{codepanel}

\section{Case Studies}

In the execution traces below, \texttt{Correct: yes/no} indicates whether the agent answered the task correctly.

\subsection{Code vs JSON Interaction}

To intuitively illustrate the structural advantages of programmatic interaction (CodeAct) compared to the traditional JSON tool invocation paradigm, we compare GPT-5 execution trajectories under both paradigms when processing the complex multi-file task in Question 1. This task requires the agent to process a total of 182 NDVI and LST files from 2019 to 2022 to calculate the interannual linear trend of the Temperature Vegetation Dryness Index (TVDI).

\textbf{(1) JSON Interaction: Context Explosion and Logic Collapse.} Under the JSON paradigm, the agent must output an independent JSON string for every tool invocation. Lacking programmatic control flows (such as \texttt{\detokenize{for}} loops), the agent attempts to manually process massive datasets, leading to severe execution inefficiency and logical errors.

\begin{codepanel}[frametitle={Case 1 --- JSON Execution Trace}]
\begin{lstlisting}[frame=none,aboveskip=0pt,belowskip=0pt,xleftmargin=0pt,xrightmargin=0pt]
[Explore]  filelist("benchmark/data/question1")
-> sees: 182 files of NDVI and LST (2019-2022)
[Explore]  skill("index"), skill("analysis"), skill("statistics")
-> identifies tools: compute_tvdi, calc_batch_image_mean, compute_linear_trend
[Explore]  doc("index.compute_tvdi")
-> learns required parameters
[Failure 1: Cherry-picking Data]
-> manually hardcodes paths for a few summer dates (July-Sept)
-> calls compute_tvdi 16 times individually instead of looping
-> completely misses the full annual representation
[Action]   calc_batch_image_mean
-> calculates spatial means for the 15 successful TVDI files (contains NaNs)
[Failure 2: Logic Error in Aggregation]
-> computes a single "global mean" (0.6919) for all dates
-> incorrectly duplicates this single value for 5 years: y=[0.6919, 0.6919, 0.6919, 0.6919, 0.6919]
[Action]   compute_linear_trend
-> calculates trend on flat line, returns slope = 0
[Failure 3: Inefficiency & Turn Limit]
-> realizes mistake, tries to recalculate annual means manually one by one
-> gets lost in indices, wastes tokens
-> hits the maximum turn limit without outputting a valid answer
Correct: no
\end{lstlisting}
\end{codepanel}

\textbf{(2) CodeAct Interaction: Efficient Programmatic Scheduling.} When using the CodeAct paradigm, the agent effectively offloads the heavy data manipulation burden to the Python environment, solving the task within a small number of turns.

\begin{codepanel}[frametitle={Case 1 --- CodeAct Execution Trace}]
\begin{lstlisting}[frame=none,aboveskip=0pt,belowskip=0pt,xleftmargin=0pt,xrightmargin=0pt]
[Explore]  Python code: filelist("benchmark/data/question1")
-> sees: 182 files, recognizes YYYY-MM-DD pattern
[Action]   Python code: parse dates and query skills
-> hits minor syntax error (.continue), corrects it immediately
[Success 1: Comprehensive Data Parsing]
-> uses Regex to dynamically match 91 valid NDVI-LST pairs across 2019-2022
-> retrieves all available data, avoiding cherry-picking
[Explore]  doc() for compute_tvdi, calc_batch_image_mean, compute_linear_trend
-> identifies inputs and outputs
[Success 2: Efficient Batch Execution]
-> writes a unified Python script to handle the entire workflow
-> loop 1: batch calculates TVDI for all 91 pairs
-> loop 2: batch calculates spatial means, utilizing try/except and np.isnan to elegantly filter invalid data
[Success 3: Spatio-temporal Aggregation]
-> accurately groups the 79 valid means by year (key slicing date_key[:4])
-> calculates true annual means for 2019, 2020, 2021, and 2022
[Action]   compute_linear_trend
-> aligns y=annual_means with x=[2019, 2020, 2021, 2022]
-> receives accurate slope: -0.037831
[Success 4: Answer Mapping]
-> matches slope (-0.037) to "Decreasing dryness at 0.037 per year"
-> outputs exact requested format B
Correct: yes
\end{lstlisting}
\end{codepanel}

\subsection{Hindsight Hint Analysis}

To intuitively demonstrate how the Hindsight Hint bridges the exploration gap in complex remote sensing workflows, we take Task 39 (extreme high-temperature monitoring) as an example and compare the teacher agent's execution trajectories with and without hint guidance.

\textbf{Task Description (Question 39):} "Using MODIS Day brightness temperature and emissivity Bands 31 over the southern Sahara edge during July 2023, calculate the number of days when more than 30\% of the region's pixels had daytime LST exceeding 315 K."

\textbf{(1) Without Hint Guidance.} Without hindsight guidance, the agent successfully explored available retrieval tools but failed to correctly establish the mapping relationship between task requirements and the dataset organization structure. Relying on an overly strict filename pattern matching \texttt{\detokenize{Emissivity_31}}, it mistakenly concluded that the emissivity input data was missing, thereby entirely missing the files actually stored under the \texttt{\detokenize{Emis31}} naming convention. This erroneous diagnosis led the agent to conduct repeated and invalid tool explorations among other alternative LST retrieval methods, ultimately terminating the task due to the inability to construct valid LST retrieval inputs.

\begin{codepanel}[frametitle={Case 2 --- Execution Trace without Hint}]
\begin{lstlisting}[frame=none,aboveskip=0pt,belowskip=0pt,xleftmargin=0pt,xrightmargin=0pt]
[Explore]  filelist(question39/)
               -> sees: MODIS BT_31 Day/Night files (July 2023)
[Explore]  skill("inversion")
               -> sees: MODIS LST inversion tools
[Explore]  doc("inversion.modis_day_night_lst")
               -> requires BT_day + BT_night + emissivity
               -> searches "Emissivity" files
[Failure Diagnosis]
               -> fails to recognize Emis31 naming convention
               -> incorrectly assumes emissivity data unavailable
[Explore]  doc("inversion.lst_multi_channel")
               -> requires BT32
               -> unavailable
[Failure]  Incorrect data discovery
               -> terminates without LST generation
Correct: no
\end{lstlisting}
\end{codepanel}

\textbf{(2) With Hindsight Hint.} To improve the subsequent sampling success rate for failed tasks, the hint generator conducts hindsight analysis based on the teacher agent's failed interaction trajectories, environmental feedback, and the ground-truth code trajectory to generate post-hoc instructional hints specific to the current task. These hints summarize key constraints, data dependencies, and potential error patterns during task execution, and are used to guide the teacher agent in re-exploring the task execution process:

\begin{codepanel}[frametitle={Case 2 --- Generated Hindsight Hint}]
\begin{lstlisting}[frame=none,aboveskip=0pt,belowskip=0pt,xleftmargin=0pt,xrightmargin=0pt]
**Data Reconnaissance & Sequential Control:** First, call the file listing tool to retrieve the actual file list. Do NOT rely on dynamic dictionary keys for iteration. You must strictly set up an integer loop using for day in range(1, 32): to iterate through all days of July 2023.
**Safe String Filtering (Critical Pitfall):** ABSOLUTELY DO NOT hardcode or filter by specific hour/minute strings (like '0850' or '0105'), as satellite pass times change daily! Inside your daily loop, build three distinct lists using basic string inclusion...
**Emissivity Stratification & Extraction:** Check if all three daily lists are non-empty. If so, apply sorted() to your daily Emissivity list. Since satellites capture night data before day data chronologically, strictly select the first file (index [0]) as nighttime and the last file (index [-1]) as daytime emissivity...
**LST Inversion:** Search the inversion kit for an LST tool explicitly designed to process dimensionally asymmetric day/night dual inputs from MODIS. Map your extracted 4 paths strictly to the tool's expected parameters...
**Spatial Ratio & Threshold Statistics:** Retrieve a statistical tool from the statistics kit capable of conditionally counting images in a batch. Pass your entire LST sequence, set the temperature threshold to 315.0, the area ratio threshold to 30.0, and the condition mode to 'above'.
**Answer Selection:** Use a dictionary mapping logic to match the returned hot day count against the specific provided options to determine the final answer.
\end{lstlisting}
\end{codepanel}

Upon receiving this hint, the teacher model successfully reconstructed the daily input tuples, selected the correct day/night LST retrieval tool, and completed the downstream threshold analysis. This example demonstrates that the proposed hint-guided data construction method can transform unsuccessful exploration trajectories into informative supervision signals, thereby improving trajectory quality while avoiding the high costs of directly generating expert trajectories.

\begin{codepanel}[frametitle={Case 2 --- Execution Trace with Hint}]
\begin{lstlisting}[frame=none,aboveskip=0pt,belowskip=0pt,xleftmargin=0pt,xrightmargin=0pt]
[Explore]  filelist(question39/)
               -> finds: BT31 Day/Night + Emis31 files

[Execute]  daily grouping (July 2023)
               -> 31 valid day/night/emissivity pairs

[Explore]  inversion.modis_day_night_lst
               -> requires 4 MODIS inputs

[Execute]  LST inversion x31
               -> generates 31 daily LST maps

[Execute]  statistics.count_images_exceeding_threshold_ratio
               -> threshold=315 K, ratio=30%
               -> 21 hot days

Correct: yes
\end{lstlisting}
\end{codepanel}

\subsection{Evolution of Policy Robustness and Efficiency}

To demonstrate the progressive capability enhancement of the agent throughout our training pipeline, we compare the execution trajectories of Qwen3-4B-Instruct and its SFT-trained and RL-optimized variants on Task 31. This case intuitively highlights the model's evolution from fragile execution to resilient error recovery (SFT), and ultimately to optimal planning (RL).

\textbf{Task Description (Question 31):} "Calculate the land surface temperature (LST) over the Taklamakan Desert near Hotan on February 23, 2020 using the split-window algorithm based on the following local input: Thermal band 31, Thermal band 32, Emissivity for band 31, Emissivity for band 32. Calculate the average surface temperature across this region."

\textbf{(1) Untrained Model: Fragile Execution and Hallucination.} The untrained agent initially successfully locates the input files and completes the LST retrieval. However, during the downstream statistical analysis phase, its performance completely collapses. It not only fabricates (hallucinates) non-existent tools and misinterprets tool purposes, but also fails to handle basic Python type mismatch errors. Unable to self-correct based on environmental feedback, the agent ultimately abandons the calculation and relies directly on its internal parametric knowledge (e.g., desert surface temperatures are typically around 290--300 K) to force a blind guess to wrap up the task.

\begin{codepanel}[frametitle={Case 3 --- Base Agent Execution Trace}]
\begin{lstlisting}[frame=none,aboveskip=0pt,belowskip=0pt,xleftmargin=0pt,xrightmargin=0pt]
[Explore]  filelist("benchmark/data/question31/")
-> Successfully found all 4 files: BT31, BT32, Emis31, Emis32.
[Explore]  skill("inversion") -> doc("inversion", "split_window")
-> Successfully mastered LST retrieval tool usage.
[Action]   Executed split_window to retrieve LST.
-> Retrieval successful, generated lst_20200223.tif.
[Failure 1: Blind Guessing Tool Name]
-> Directly called statistics.calculate_mean_value in code to compute the mean.
-> Error: ValueError, the tool does not exist at all!
[Failure 2: Misunderstanding Tool Purpose]
-> Glanced at the available tool list and randomly picked calculate_tif_average.
-> Error: Missing output_path. This tool is designed to "composite multiple TIFs into an average TIF," not to calculate the numerical average of a single image.
-> Forced an output_path and ran it again; returned a file path, failed to extract a number.
[Failure 3: Type Mismatch]
-> Randomly switched tools again, attempting to use statistics.mean.
-> Error: TypeError. The mean tool expects a list of numbers (list[float]), but the agent passed in an image file path.
[Failure 4: Complete Breakdown & Hallucination]
-> Completely deadlocked, abandons calculation. Based on the common sense that "desert surface temperature is usually 280--300 K," the agent forces option B (294.65 K) as the answer.
Correct: no (Unable to recover from errors based on feedback)
\end{lstlisting}
\end{codepanel}

\begin{table*}[!t]
\centering
\caption{Comprehensive Performance Evaluation Results. All metrics except Efficiency are reported as percentages.}
\label{tab:table8}
\scriptsize
\resizebox{\linewidth}{!}{%
\begin{tabular}{llcccccccccccc}
\toprule
\multirow{2}{*}{\textbf{Model}} & \multirow{2}{*}{\textbf{Paradigm}} & \multicolumn{2}{c}{\textbf{Accuracy}} & \multicolumn{2}{c}{\textbf{Efficiency}} & \multicolumn{2}{c}{\textbf{Tool-Any-Order}} & \multicolumn{2}{c}{\textbf{Tool-In-Order}} & \multicolumn{2}{c}{\textbf{Tool-Exact-Match}} & \multicolumn{2}{c}{\textbf{Parameters}} \\
\cmidrule(lr){3-4} \cmidrule(lr){5-6} \cmidrule(lr){7-8} \cmidrule(lr){9-10} \cmidrule(lr){11-12} \cmidrule(lr){13-14}
& & \textbf{AP} & \textbf{IF} & \textbf{AP} & \textbf{IF} & \textbf{AP} & \textbf{IF} & \textbf{AP} & \textbf{IF} & \textbf{AP} & \textbf{IF} & \textbf{AP} & \textbf{IF} \\
\midrule
Qwen3-4B-Instruct & Code & 31.8 & 35.8 & 3.030 & 3.904 & 47.6 & 52.8 & 40.1 & 45.8 & 35.4 & 37.8 & 23.8 & 24.2 \\
Qwen3-4B SFT & Code & 41.5 & 44.3 & 4.650 & 4.356 & 58.6 & 60.9 & 49.4 & 51.8 & 42.9 & 42.6 & 23.9 & 24.8 \\
Qwen3-4B RL (Ours) & Code & 65.9 & 65.9 & 4.917 & 6.446 & 69.1 & 70.1 & 58.9 & 58.4 & 48.6 & 48.2 & 24.1 & 24.0 \\
Qwen3-8B & Code & 19.3 & 22.7 & 3.957 & 3.907 & 49.6 & 50.9 & 41.1 & 44.2 & 36.9 & 37.8 & 23.8 & 24.8 \\
Qwen3-8B SFT & Code & 46.6 & 44.3 & 4.242 & 5.158 & 61.4 & 63.6 & 52.2 & 53.2 & 44.3 & 42.7 & 25.3 & 25.1 \\
Qwen3-8B RL (Ours) & Code & 61.4 & 64.2 & 4.180 & 4.794 & 67.8 & 70.5 & 57.8 & 60.0 & 48.5 & 48.5 & 25.7 & 25.7 \\
Qwen3-32B & Code & 43.8 & 42.6 & 5.307 & 4.532 & 59.3 & 60.3 & 47.8 & 50.7 & 41.6 & 42.6 & 18.0 & 14.0 \\
DeepSeek-V3.1 & Code & 60.8 & 65.3 & 10.401 & 11.065 & 71.2 & 73.1 & 60.0 & 62.4 & 41.3 & 46.0 & 25.5 & 26.0 \\
GPT-5 & Code & 71.6 & 71.0 & 7.763 & 9.468 & 69.1 & 72.0 & 58.4 & 61.8 & 48.2 & 49.9 & 18.3 & 17.2 \\
Qwen3-32B & JSON & 35.8 & 34.1 & 1.556 & 2.097 & 54.1 & 63.4 & 36.7 & 50.0 & 7.5 & 21.7 & 6.0 & 12.0 \\
DeepSeek-V3.1 & JSON & 55.7 & 55.1 & 2.975 & 3.004 & 78.7 & 79.6 & 65.0 & 66.1 & 48.7 & 50.1 & 25.6 & 26.0 \\
GPT-5 & JSON & 68.8 & 71.6 & 3.817 & 4.266 & 72.4 & 75.2 & 58.5 & 61.0 & 47.7 & 46.3 & 22.6 & 22.6 \\
\bottomrule
\end{tabular}}
\end{table*}

\textbf{(2) SFT Model: Trial-and-Error and Correction.} After Masked SFT training, the agent could interpret environmental feedback and debug its code. Its strict date filter failed because BT filenames use underscores whereas Emis filenames use hyphens. Through repeated Python debugging, it identified the correct paths and completed the task in 11 turns, demonstrating error recovery but requiring a relatively long trajectory.

\begin{codepanel}[frametitle={Case 3 --- SFT Agent Execution Trace}]
\begin{lstlisting}[frame=none,aboveskip=0pt,belowskip=0pt,xleftmargin=0pt,xrightmargin=0pt]
[Explore]  Python: filelist("benchmark/data/question31")
-> Discovered there are only 4 relevant files in total.
[Failure 1: Over-engineered Filtering]
-> Attempted to filter all files using a strict date format with underscores: target_date = "2020_02_23".
-> Pitfall: Emis files are named using hyphens "2020-02-23", causing the Emis path extraction to fail (returned None).
[Action]   Python Debugging (Multiple Trial-and-Error)
-> Re-searched for files containing only "Emis31".
-> Triggered a system format error midway, modified the code to only print the number of files.
-> Finally confirmed file existence and extracted the complete Emis path step-by-step.
[Explore]  skill("inversion") -> doc("inversion", "split_window")
-> Mastered LST retrieval tool usage and parameters.
[Action]   Python: Passed the finally gathered 4 paths to invoke split_window.
-> Successfully generated LST image.
[Explore]  skill("statistics") -> doc("statistics", "calc_batch_image_mean")
-> Accurately found and mastered the image mean calculation tool.
[Action]   Python: calc_batch_image_mean([lst_output_path])
-> Successfully extracted the mean value 301.2203 K.
[Success: Answer Mapping]
-> Matched option C.
Correct: yes (Recovered from errors based on feedback)
Total: 11 Turns
\end{lstlisting}
\end{codepanel}

\textbf{(3) RL Model: Minimalist and Efficient Execution.} Under the joint optimization of process rewards and efficiency penalties, the RL agent exhibited highly robust and minimalist reasoning capabilities. It completely bypassed the fragile date string matching trap, directly adopting a core keyword matching strategy. Furthermore, it seamlessly consolidated file extraction and tool invocation into a single code block for execution, drastically reducing interaction overhead and perfectly solving the task in 8 turns.

\begin{codepanel}[frametitle={Case 3 --- RL Agent Execution Trace}]
\begin{lstlisting}[frame=none,aboveskip=0pt,belowskip=0pt,xleftmargin=0pt,xrightmargin=0pt]
[Explore]  Python: filelist("benchmark/data/question31")
-> Discovered there are only 4 relevant files in total.
[Explore]  skill("inversion") -> doc("inversion", "split_window")
-> Mastered LST retrieval tool usage and parameters.
[Success 1: Direct File Matching & Combined Execution]
-> Realized there are only 4 files in total, abandoned redundant date filtering, and directly matched the whole list using core keywords ("BT_31", "Emis31").
-> Wrote path extraction and the split_window tool invocation in the same code block for direct execution.
-> Successfully generated the LST image in one go.
[Explore]  skill("statistics") -> doc("statistics", "calc_batch_image_mean")
-> Accurately found and mastered the image mean calculation tool.
[Action]   Python: calc_batch_image_mean([lst_output_path])
-> Successfully extracted the mean value 301.2203 K.
[Success 2: Answer Mapping]
-> Directly matched option C.
Correct: yes
Total: 8 Turns
\end{lstlisting}
\end{codepanel}

\section{Complete Earth-Bench Evaluation Results}

This section provides the complete evaluation results of our proposed method on the Earth-Bench benchmark, including end-to-end task completion metrics, tool execution process metrics, and parameter-level matching metrics. All results are based on the 176 test tasks held out from training and are calculated under both Autonomous Planning (AP) and Instruction Following (IF) evaluation modes.

As observed from Table~\ref{tab:table8}, the \emph{Code} paradigm generally exhibits higher Efficiency values. For instance, in AP mode, the Efficiency of GPT-5 under the \emph{Code} paradigm is 7.763, compared with 3.817 under the \emph{JSON} paradigm. Under the definition of this metric, these higher values indicate more tool invocations relative to the reference trajectory.

Both interaction paradigms are evaluated under a fixed budget of at most 25 assistant turns per task. In the \emph{JSON} paradigm, tool invocations are issued through explicit structured calls, whereas a single \emph{Code} turn can execute multiple tool operations through loops or batch processing. Grouping calls within a turn does not itself increase the call count, but it allows multiple calls to be repeated within a retry turn. When a code block fails after partially completing a sequence of operations, retrying that block may re-execute calls that already succeeded, increasing the cumulative number of tool invocations within the same turn budget.

Earth-Bench counts the tool invocations executed before an exception as well as those performed during subsequent retries. Repeated execution can therefore contribute to the higher call-count ratios observed under the \emph{Code} paradigm. These ratios should be distinguished from token overhead and the number of assistant turns: more tool calls relative to the reference do not necessarily imply more action or observation tokens, or more interaction turns.

The Tool-Exact-Match and Parameters metrics require the agent to strictly reproduce the tool sequence and parameters in the reference trajectory, making them more sensitive to differences in execution paths compared to Tool-Any-Order and Tool-In-Order.

The complete results show that even if a model has a high Accuracy, its Tool-Exact-Match is still generally lower than its Tool-In-Order. This indicates that multiple valid execution paths exist for remote sensing tasks, and successfully completing a task does not require strictly replicating the reference tool sequence. Especially under the \emph{Code} paradigm, the agent can achieve equivalent computational processes through program logic, intermediate variable reuse, and different tool combinations. Consequently, the predicted trajectory may have formal differences from the reference trajectory.

\end{document}